\documentclass[runningheads]{llncs}
\usepackage{xfp}
\usepackage{siunitx}
\usepackage{listings}
\usepackage{datetime}
\usepackage{amsmath}
\usepackage{graphicx}
\usepackage{comment}
\usepackage{enumitem}
\usepackage{booktabs}
\usepackage{amssymb}
\usepackage{algorithm}
\usepackage{algpseudocode}
\usepackage{subcaption}
\usepackage{caption}

\usepackage{tikz}
\usepackage{pgfplots}
\usepgfplotslibrary{fillbetween}
\usetikzlibrary{pgfplots.groupplots}
\usepgfplotslibrary{groupplots}

\usepackage{xcolor}
\definecolor{codegreen}{rgb}{0,0.6,0}
\definecolor{codegray}{rgb}{0.5,0.5,0.5}
\definecolor{codepurple}{rgb}{0.58,0,0.82}
\definecolor{backcolour}{rgb}{1.0,1.0,1.0}

\lstdefinestyle{mystyle}{
	backgroundcolor=\color{backcolour},   
	commentstyle=\color{codegreen},
	keywordstyle=\color{magenta},
	numberstyle=\tiny\color{codegray},
	stringstyle=\color{codepurple},
	basicstyle=\ttfamily\scriptsize,
	breakatwhitespace=false,         
	breaklines=true,                 
	captionpos=b,                    
	keepspaces=true,                 
	numbers=left,                    
	numbersep=5pt,                  
	showspaces=false,                
	showstringspaces=false,
	showtabs=false,                  
	tabsize=1
}

\newcommand{\ra}[1]{\renewcommand{\arraystretch}{#1}}
\usepackage[abbreviations]{foreign}

\usepackage[pagebackref=true,breaklinks=true,colorlinks,bookmarks=false,citecolor=blue]{hyperref}

\newcommand{\topic}[1]{\noindent\small{\underline{\textbf{#1:}}}}

\def\softmax{\text{SM}}
\def\Ver{/c3}
\begin{document}
\def\blfootnote{\xdef\@thefnmark{}\@footnotetext}
\title{Deep is a Luxury We Don\textquotesingle t Have}
%
%

\def\final{}

\ifx\final\undefined
\author{}
\authorrunning{Anonymous~\etal}
\institute{WhiteRabbit.AI}

\else

\makeatletter
\newcommand{\printfnsymbol}[1]{%
	\textsuperscript{\@fnsymbol{#1}}%
}
\makeatother
\author{Ahmed Taha\thanks{Equal Contribution} \and 
	Yen Nhi Truong Vu\printfnsymbol{1} \and
	Brent Mombourquette  \and
	Thomas Paul Matthews \and
	Jason Su\and
	Sadanand Singh}
\authorrunning{A. Taha~\etal}
\institute{WhiteRabbit.AI}
\fi

\maketitle              

\begin{abstract}
	Medical images come in high resolutions. A high resolution is vital for finding malignant tissues at an early stage. Yet, this resolution presents a challenge in terms of modeling long range dependencies. Shallow transformers eliminate this problem, but they suffer from quadratic complexity. In this paper, we tackle this complexity by leveraging a linear self-attention approximation. Through this approximation, we propose an efficient vision model called \textbf{HCT} that stands for \textbf{H}igh resolution \textbf{C}onvolutional \textbf{T}ransformer. HCT brings  transformers' merits  to high resolution images at a significantly lower cost. We evaluate HCT using a high resolution mammography dataset. HCT is significantly superior to its CNN counterpart. Furthermore, we demonstrate HCT’s fitness for medical images by evaluating its effective receptive field. Code available at \textit{https://bit.ly/3ykBhhf}
	
	
	
	
\end{abstract}
\section{Introduction}\label{sec:introduction}
Medical images have high spatial dimensions (\eg 6 million pixels). This poses a challenge in terms of modeling long range spatial dependencies. To learn these dependencies, one can build a deeper network by stacking more layers. Yet, this fails for two reasons: (1) the computational cost of each layer is significant because of the high resolution input; (2) stacking convolutional layers expands the \textit{effective} receptive field sublinearly~\cite{luo2016understanding}. This means that a huge number of layers is needed. For high resolution inputs, a deep network is a luxury. 

Compared to convolutional neural networks (CNNs), transformers~\cite{bahdanau2014neural,vaswani2017attention} are superior in terms of modeling long range dependencies. Yet, transformers are computationally expensive due to the attention layers' quadratic complexity.  Furthermore, transformers are data hungry~\cite{dosovitskiy2020image}. This limits the utility of transformers in medical images which are both high resolution and scarce. In this paper, we tackle these challenges and propose a \textbf{H}igh resolution \textbf{C}onvolutional \textbf{T}ransformer (HCT). Through HCT, we achieve statistically significant better performance and emphasize the following message: Keep an eye on your network’s effective receptive field (ERF).


\newcommand{\scale}{0.21}

\newcommand{\hscale}{0.19} 
\newcommand{\wscale}{0.15} 

\begin{table}[t]
	\scriptsize
	\centering
	\caption{Effective receptive field~\cite{luo2016understanding} (ERF) evaluation for GMIC~\cite{shen2019globally} and HCT. The first row shows the ERF using 100 breast images (left and right) randomly sampled. The second and third rows show the ERF using 100 right and left breast images randomly sampled, respectively. To highlight the ERF difference, we aggregate the ERF across images' rows and columns. The GMIC's ERF (blue curves) is highly concentrated around the center pixel and has a Gaussian shape. In contrast, the HCT's ERF (orange curves) is less concentrated around the center and spreads \textit{dynamically} to the breasts' locations without explicit supervision. Finally, the square-root of ERF (Sqrt EFR) highlights the difference between the GMIC and HCT’s ERF. These high resolution images are best viewed on a screen.}
	\label{tab:receptive_field_analysis}
	\setlength{\tabcolsep}{2pt}
	\begin{tabular}{@{}cccccccc@{}}
		\toprule
		\multicolumn{2}{c}{ERF} &\phantom{a}& \multicolumn{2}{c}{Aggregated ERF} &\phantom{a}& \multicolumn{2}{c}{Sqrt ERF}\\
		\cmidrule{1-2} \cmidrule{4-5} \cmidrule{7-8}
		GMIC & HCT (Ours) & & Across Rows & Across Cols && GMIC & HCT (Ours) \\
		\midrule
		\includegraphics[width=\wscale\linewidth,height=\hscale\linewidth]{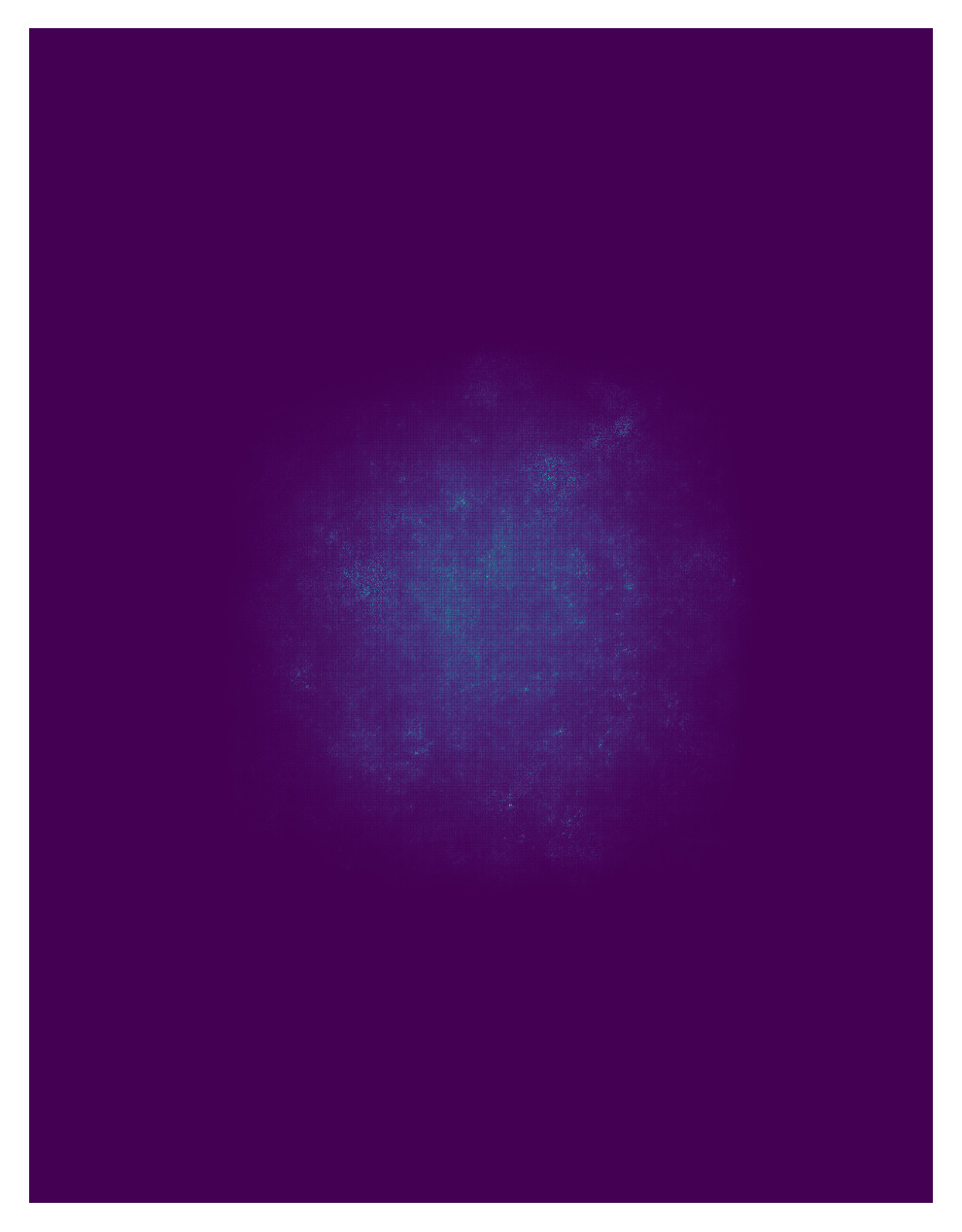} &
		\includegraphics[width=\wscale\linewidth,height=\hscale\linewidth]{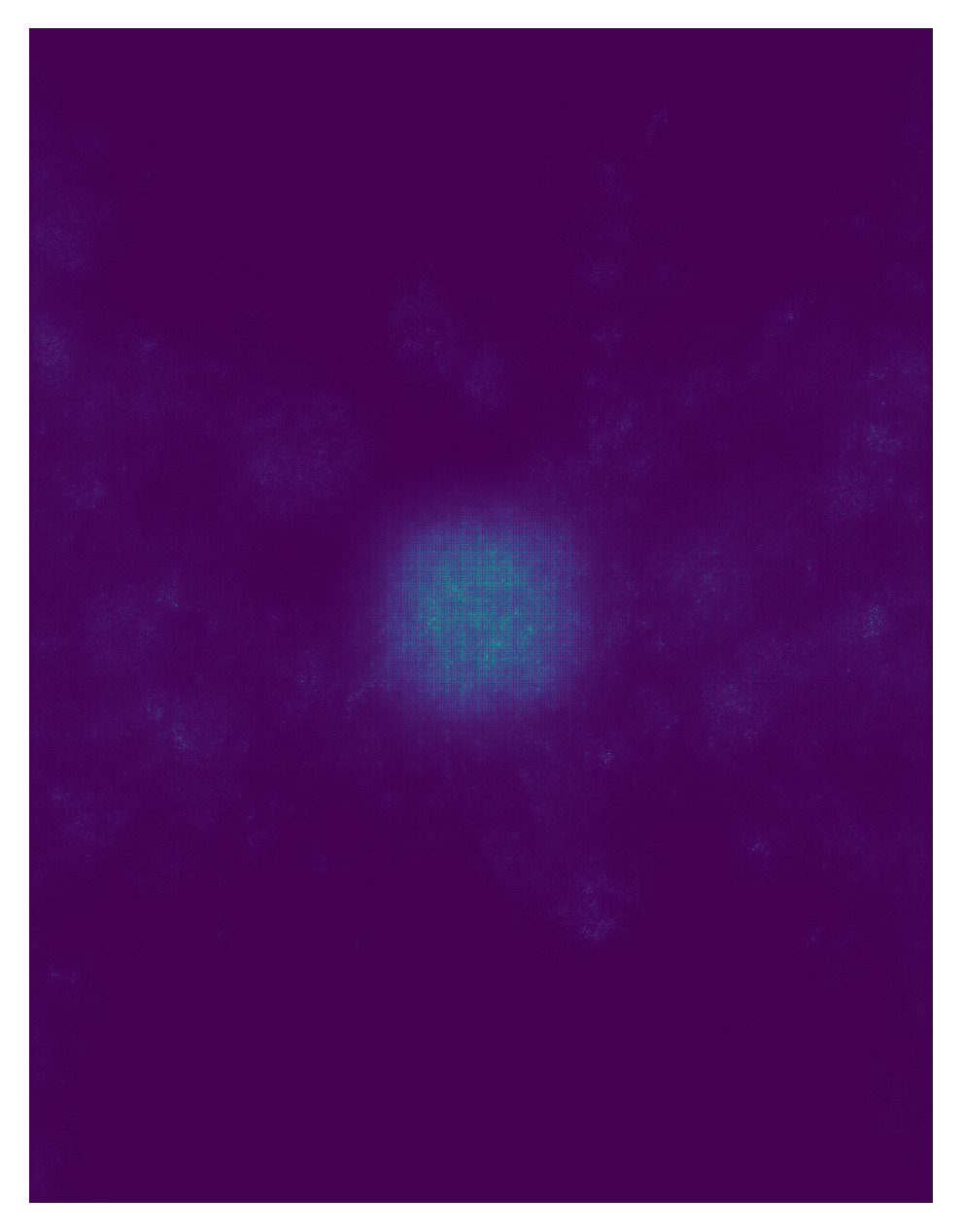} &&
		\includegraphics[width=\wscale\linewidth,height=\hscale\linewidth]{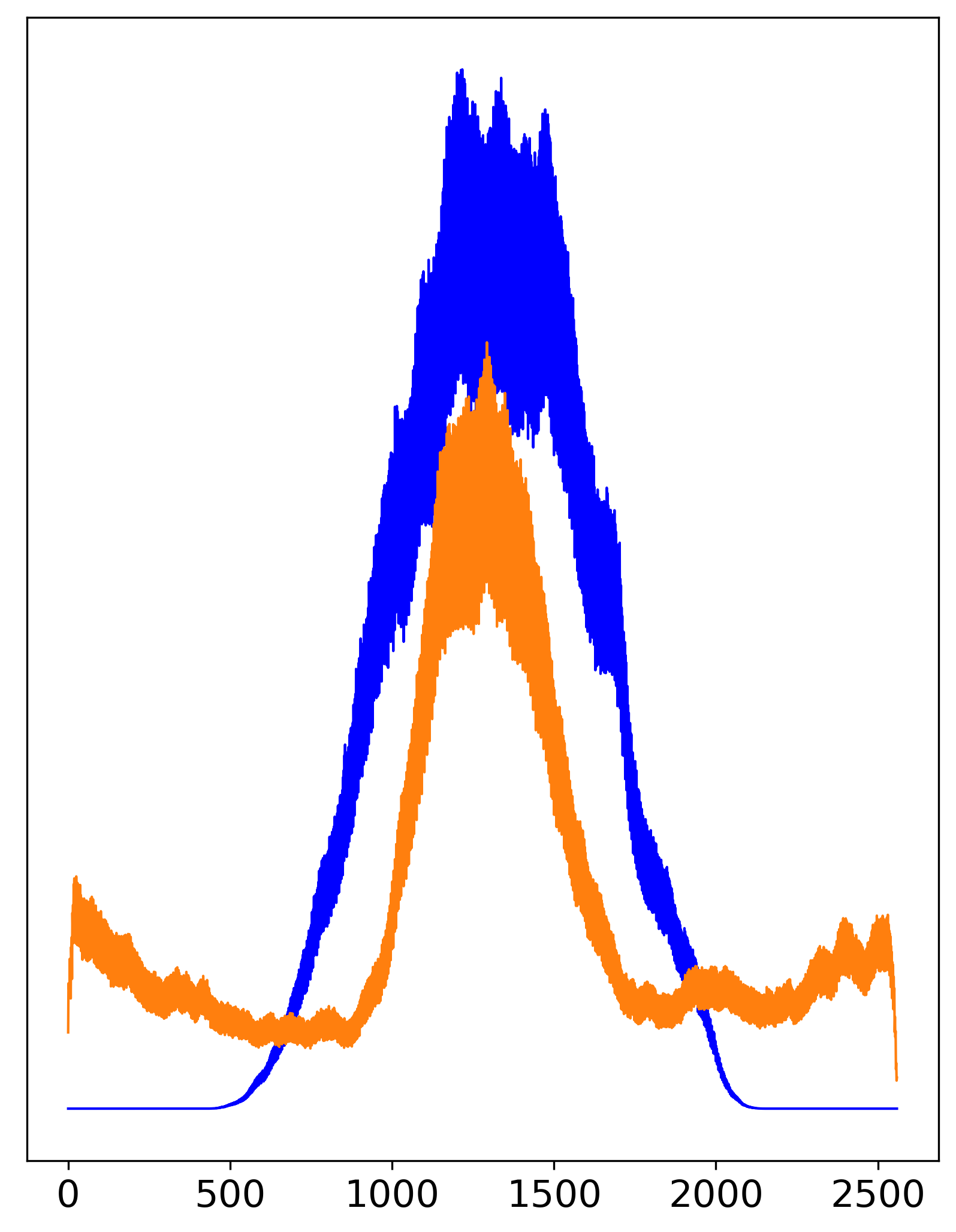} &
		\includegraphics[width=\wscale\linewidth,height=\hscale\linewidth]{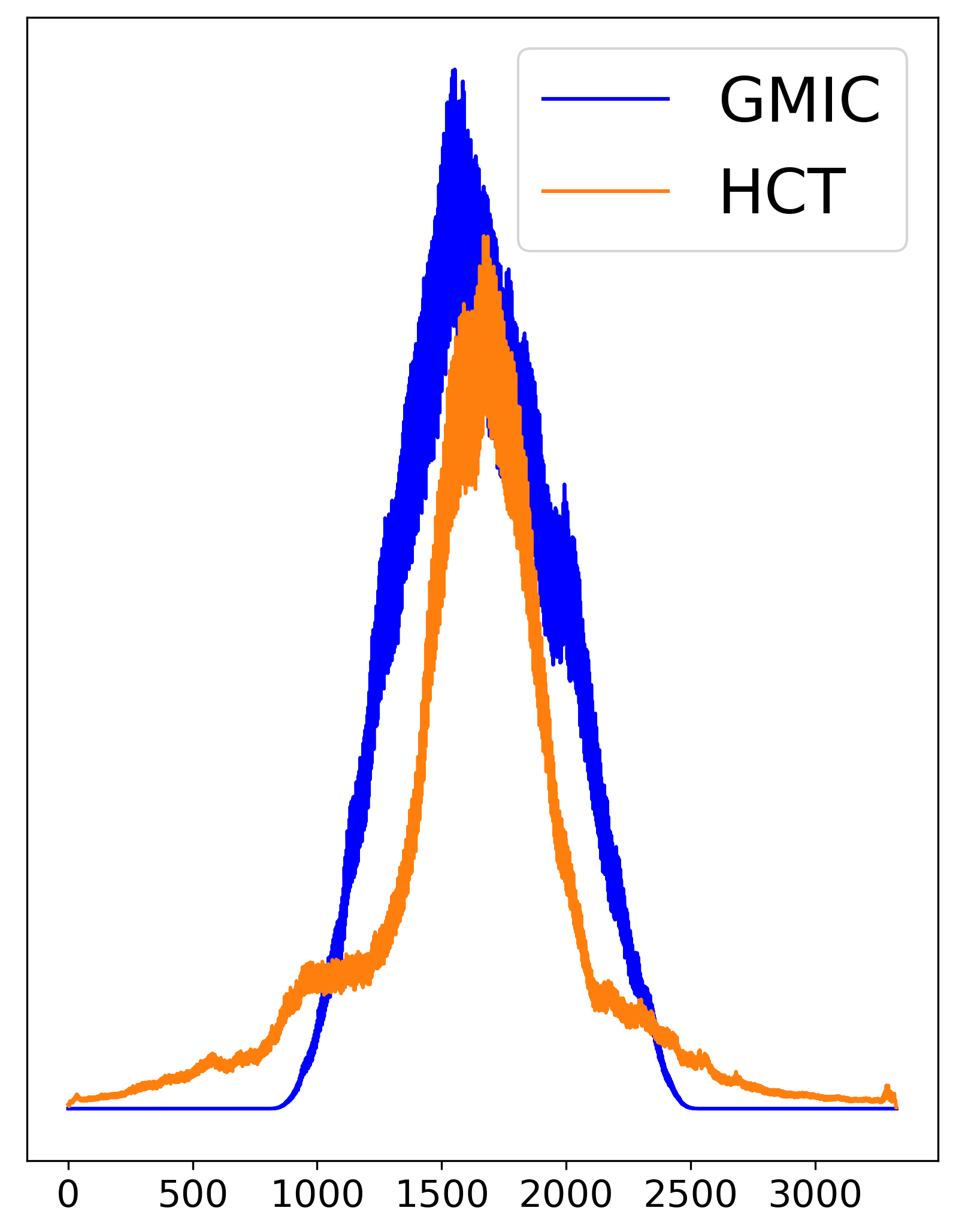} &&
		\includegraphics[width=\wscale\linewidth,height=\hscale\linewidth]{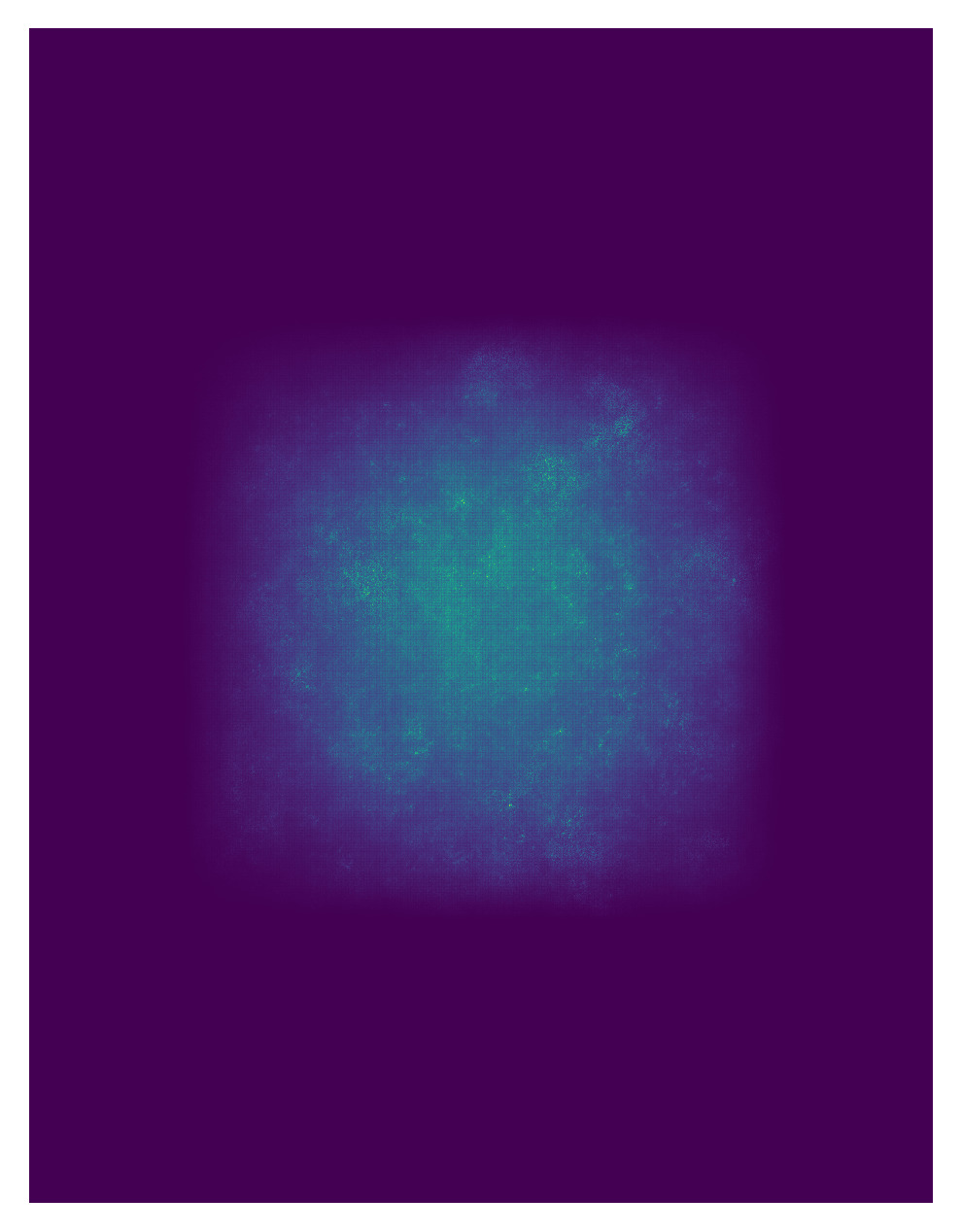} &
		\includegraphics[width=\wscale\linewidth,height=\hscale\linewidth]{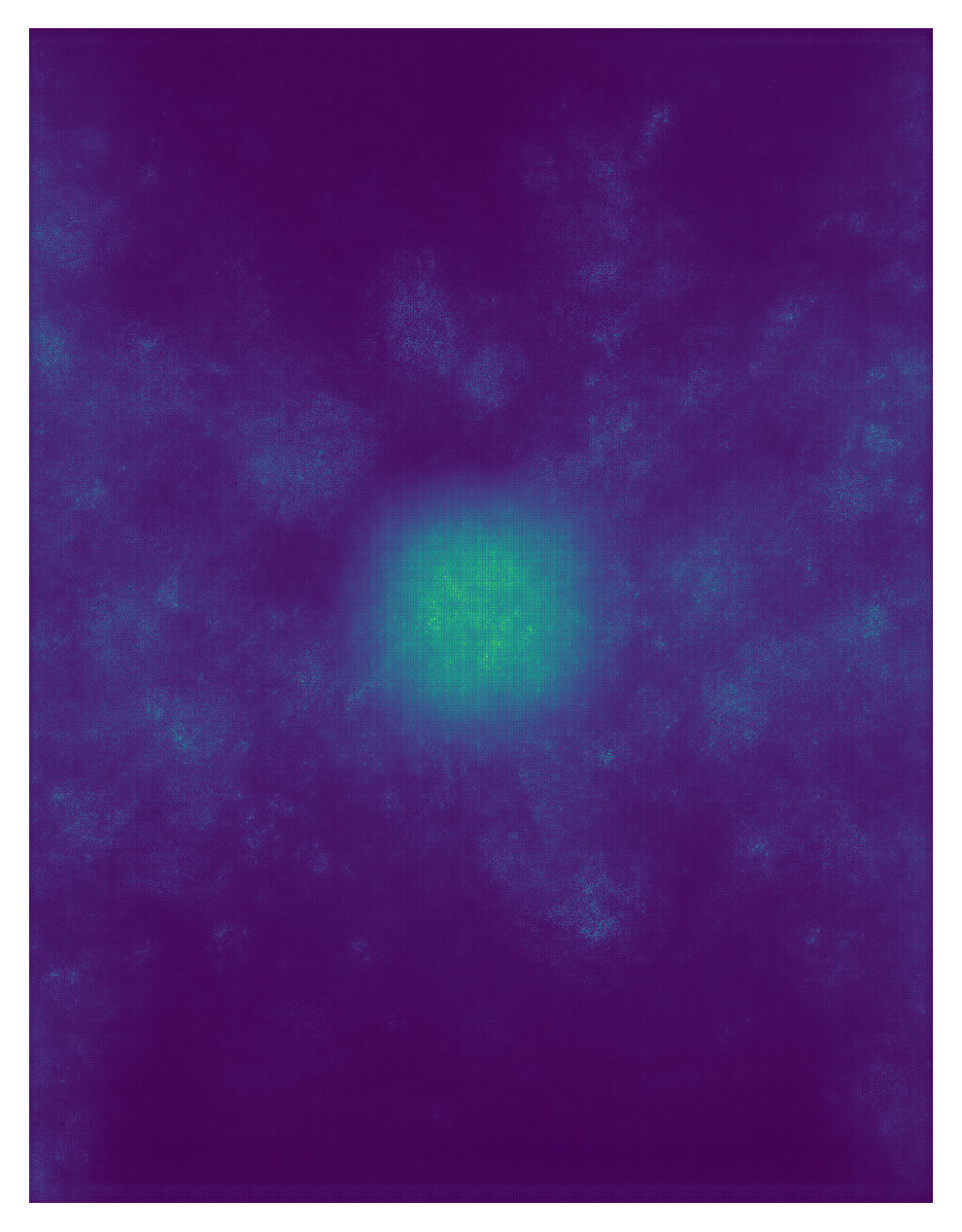} 
		\\
		\includegraphics[width=\wscale\linewidth,height=\hscale\linewidth]{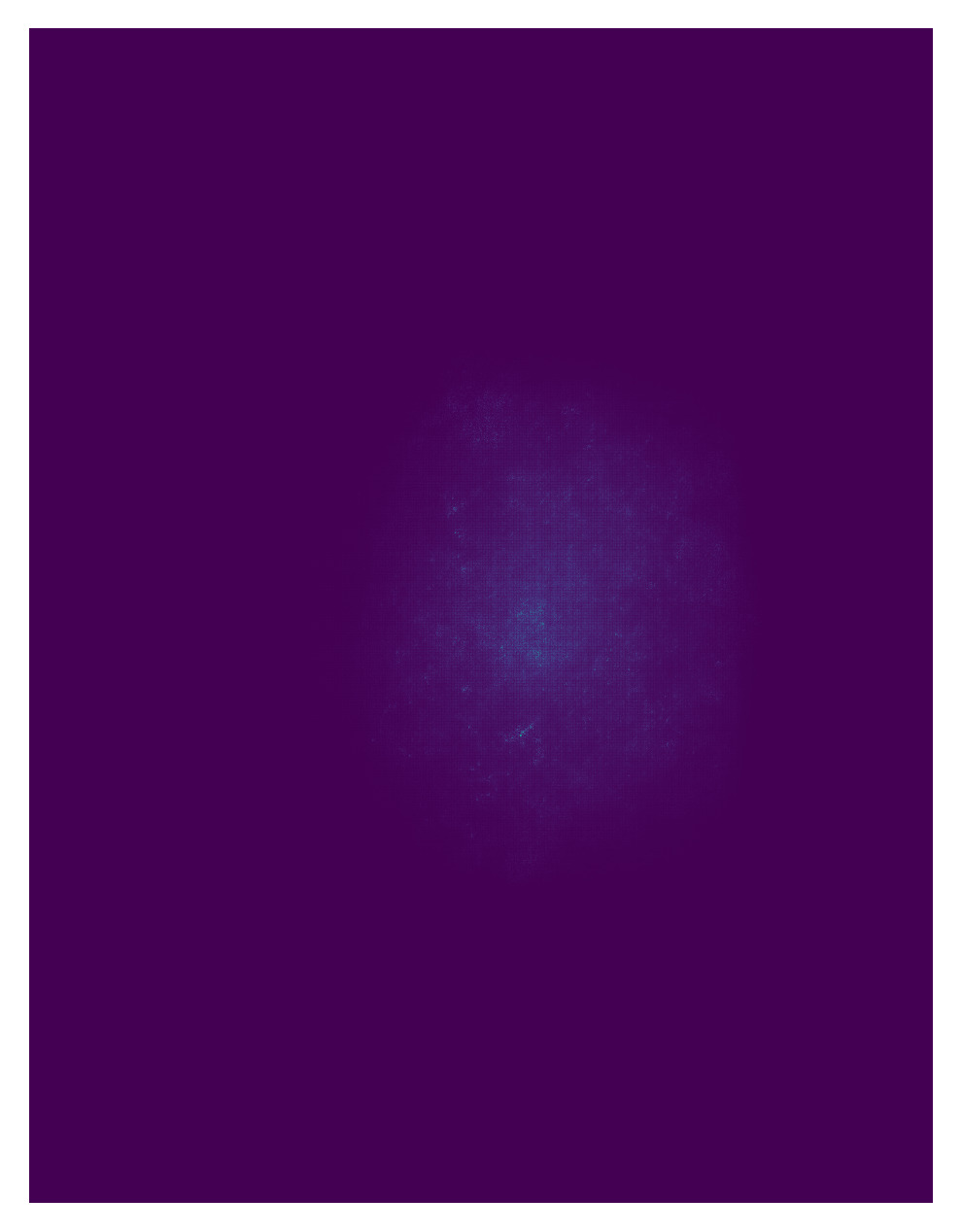} &
		\includegraphics[width=\wscale\linewidth,height=\hscale\linewidth]{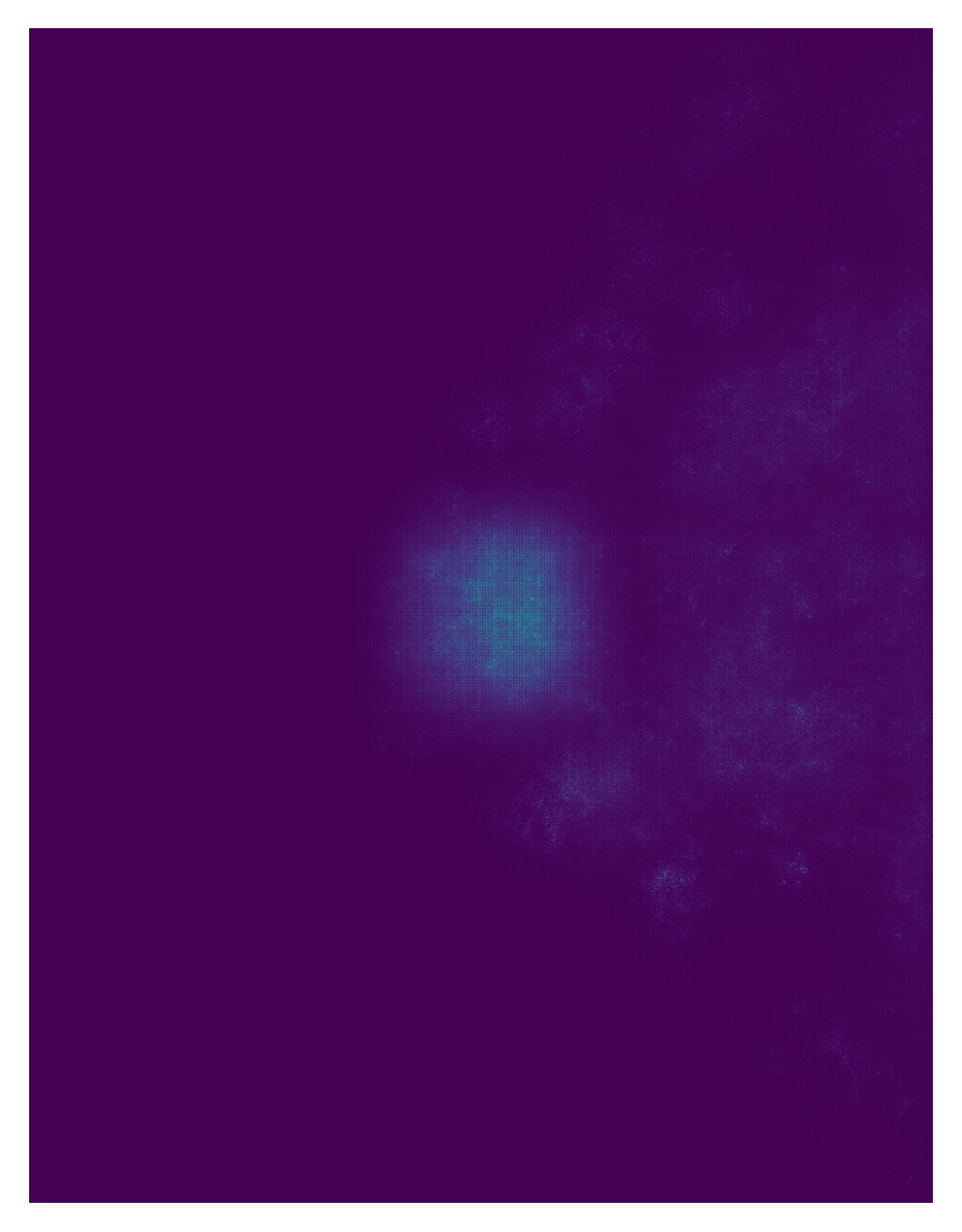} &&
		\includegraphics[width=\wscale\linewidth,height=\hscale\linewidth]{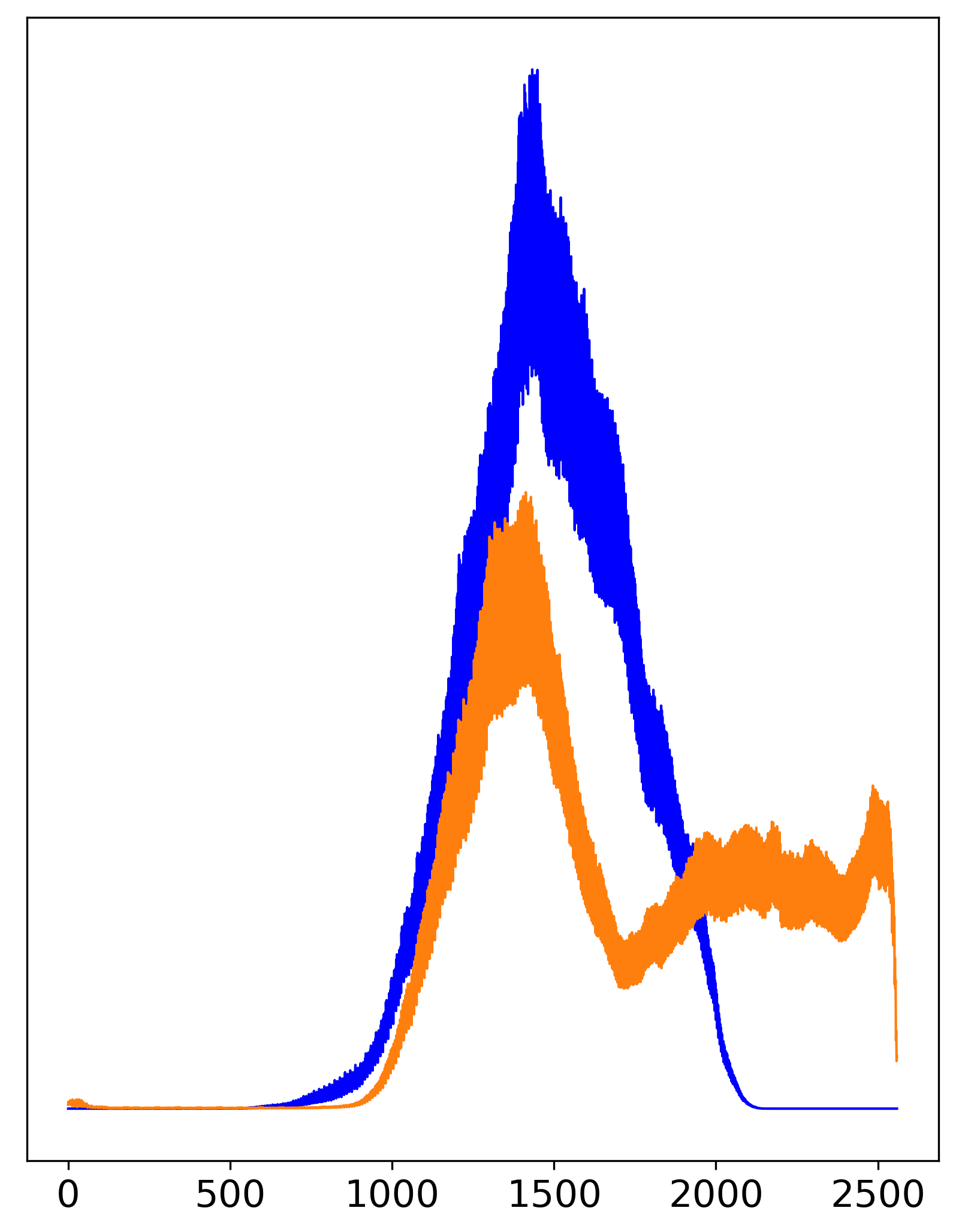} &
		\includegraphics[width=\wscale\linewidth,height=\hscale\linewidth]{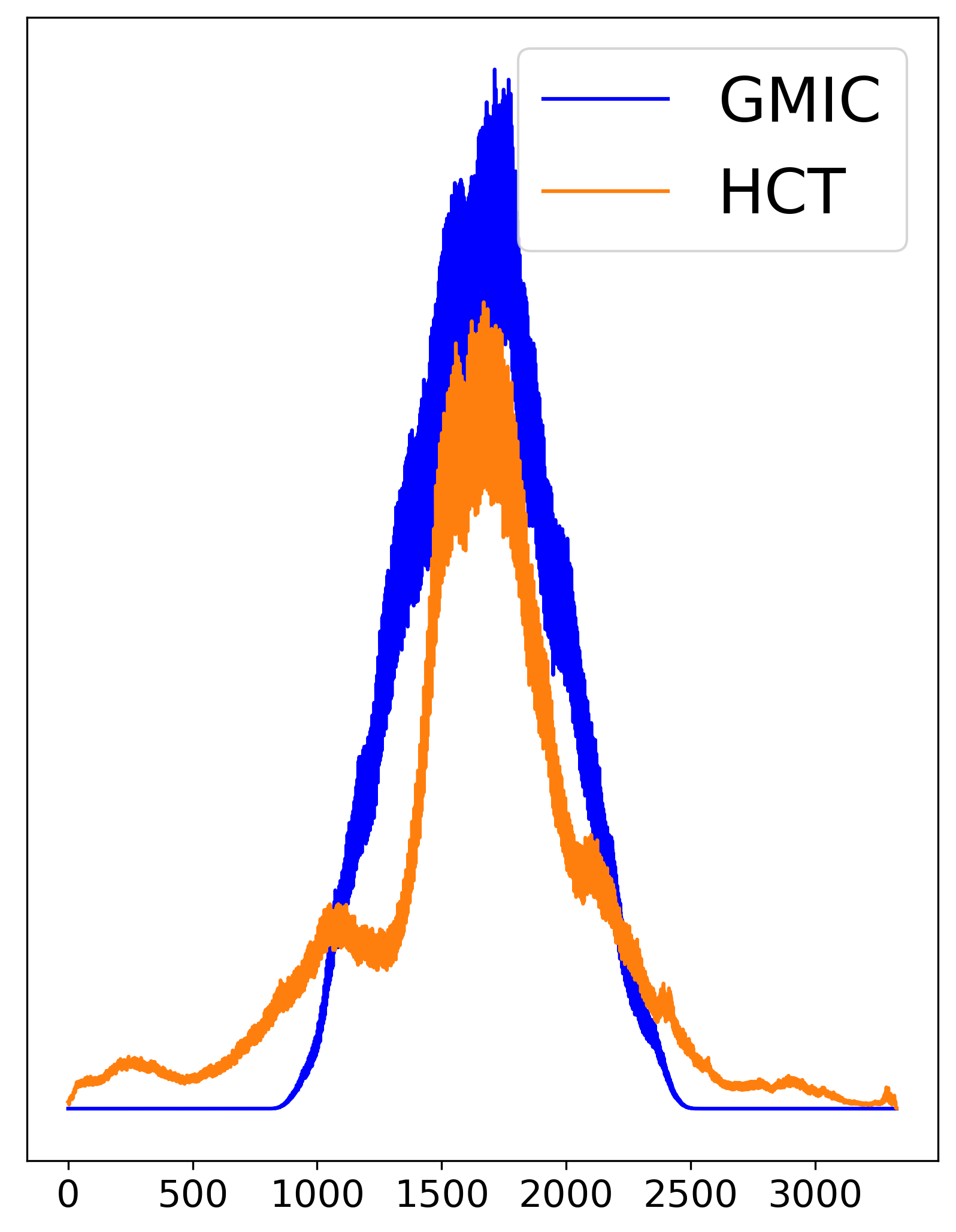} &&
		\includegraphics[width=\wscale\linewidth,height=\hscale\linewidth]{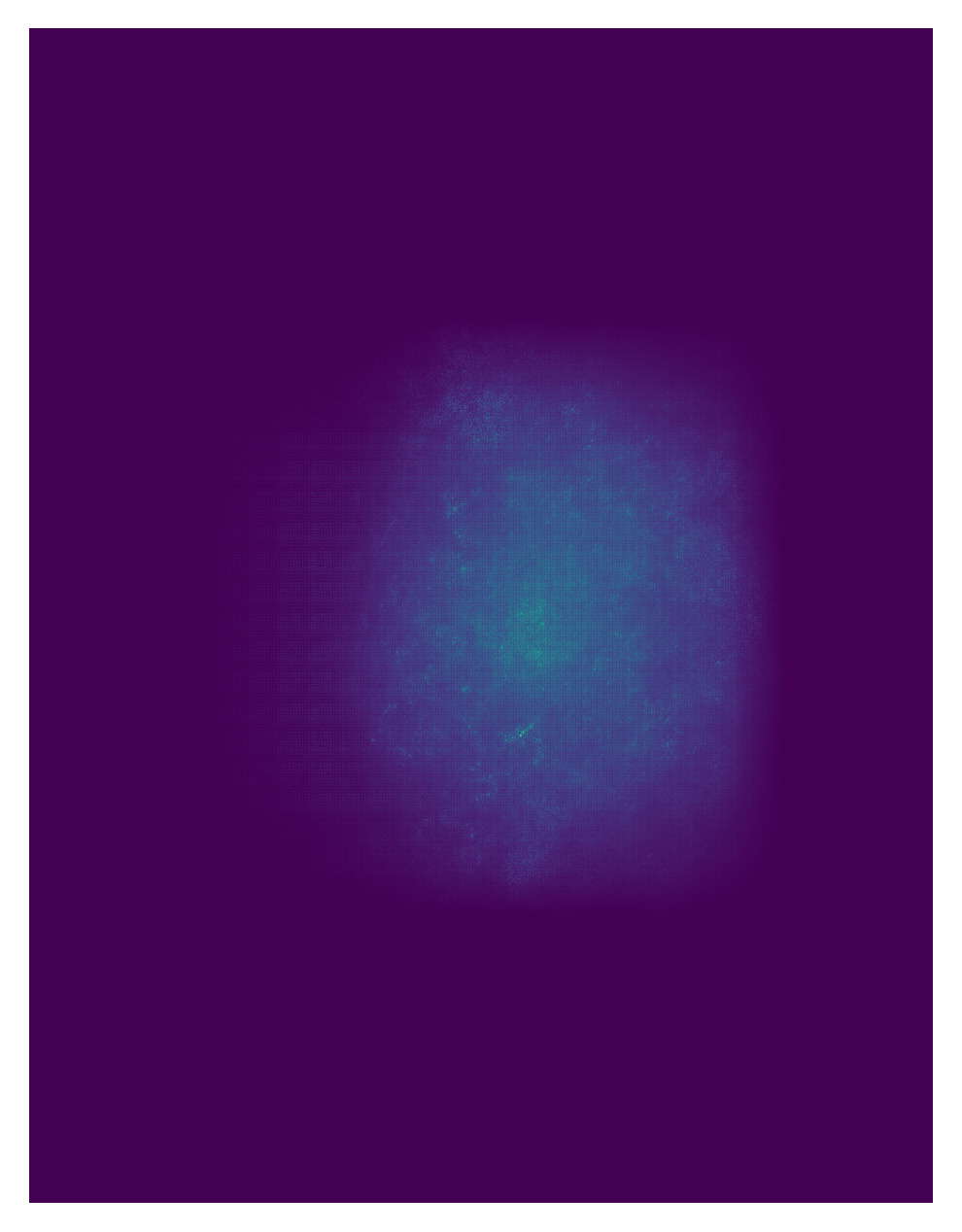} &
		\includegraphics[width=\wscale\linewidth,height=\hscale\linewidth]{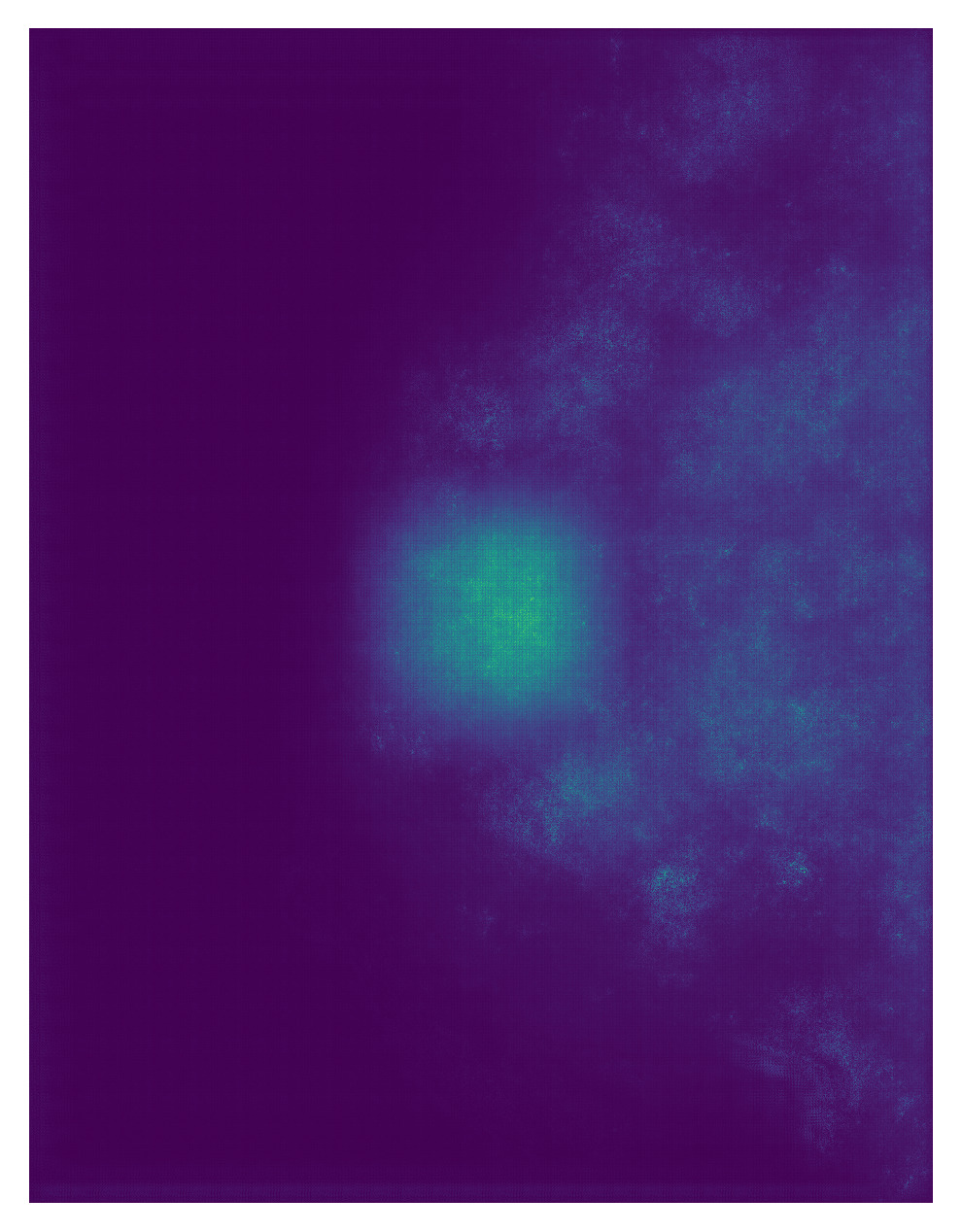} 
		\\
		\includegraphics[width=\wscale\linewidth,height=\hscale\linewidth]{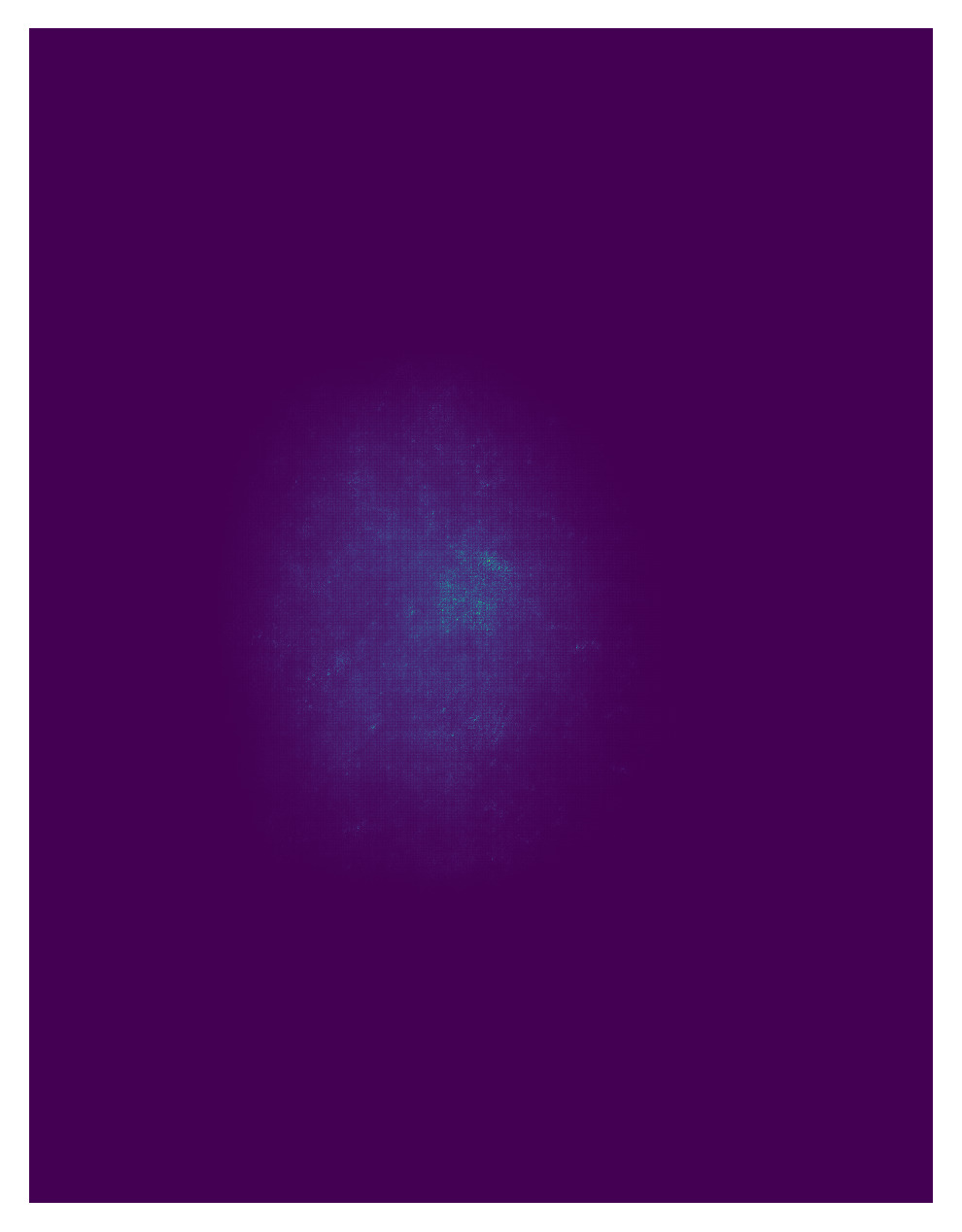} &
		\includegraphics[width=\wscale\linewidth,height=\hscale\linewidth]{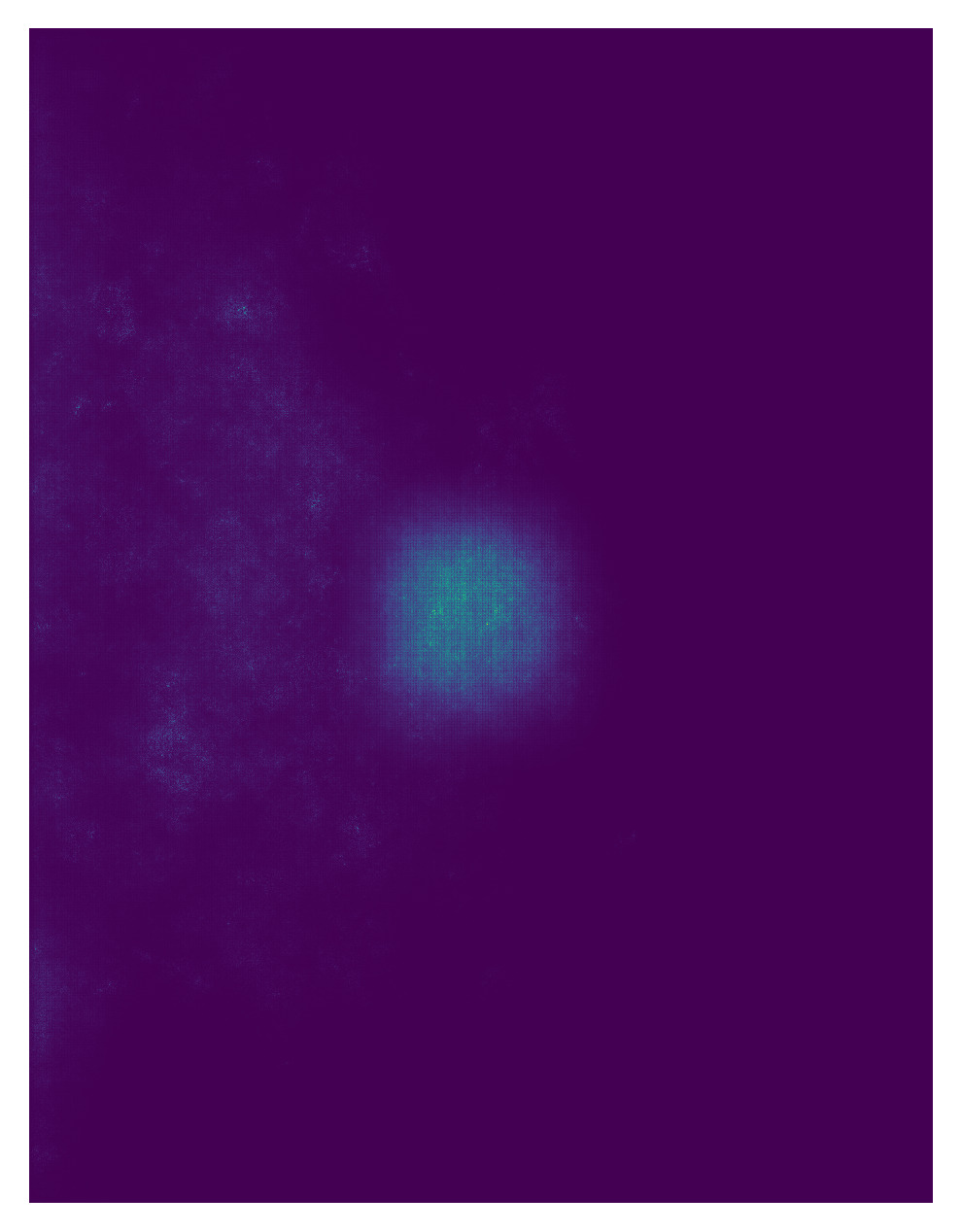} &&
		\includegraphics[width=\wscale\linewidth,height=\hscale\linewidth]{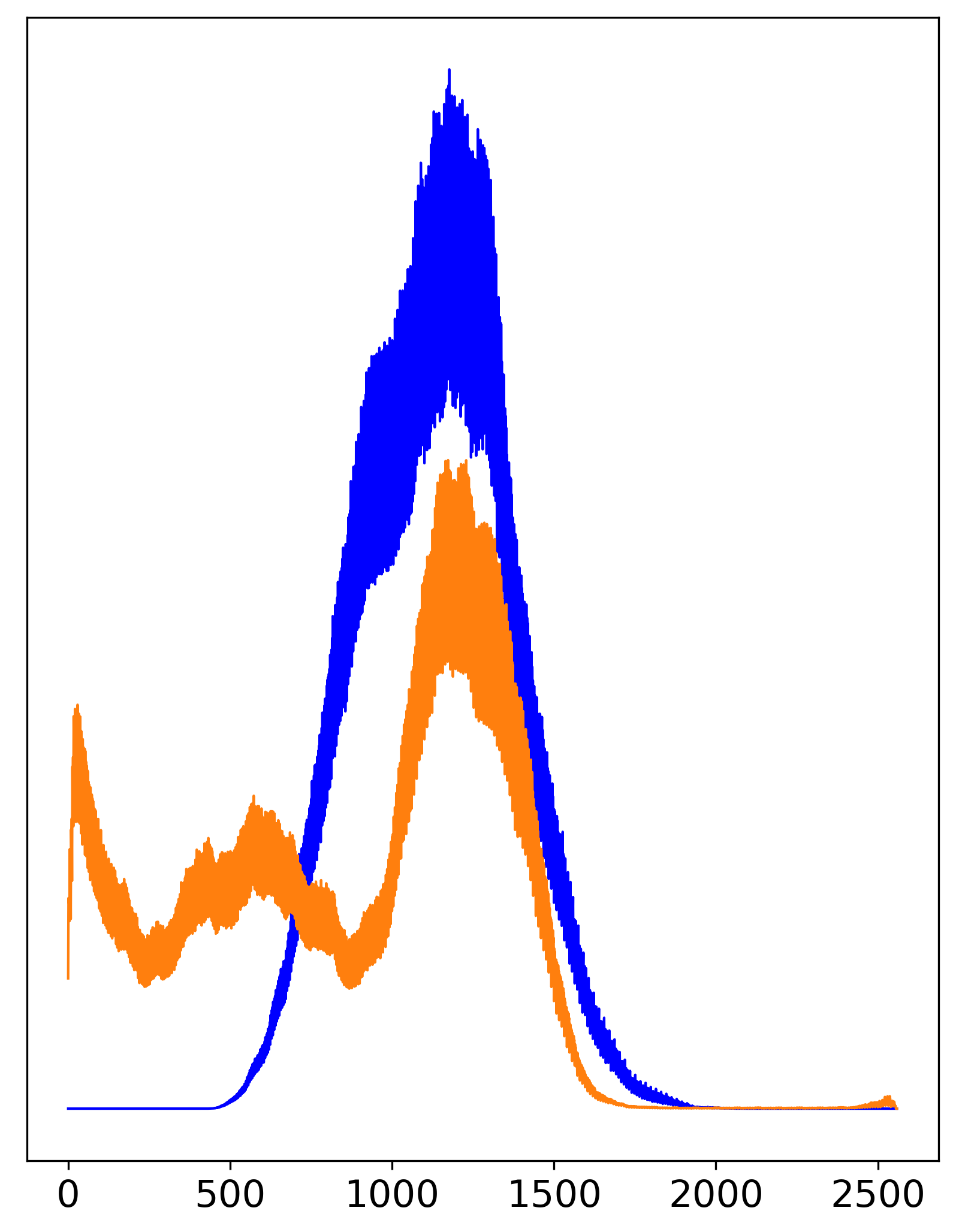} &
		\includegraphics[width=\wscale\linewidth,height=\hscale\linewidth]{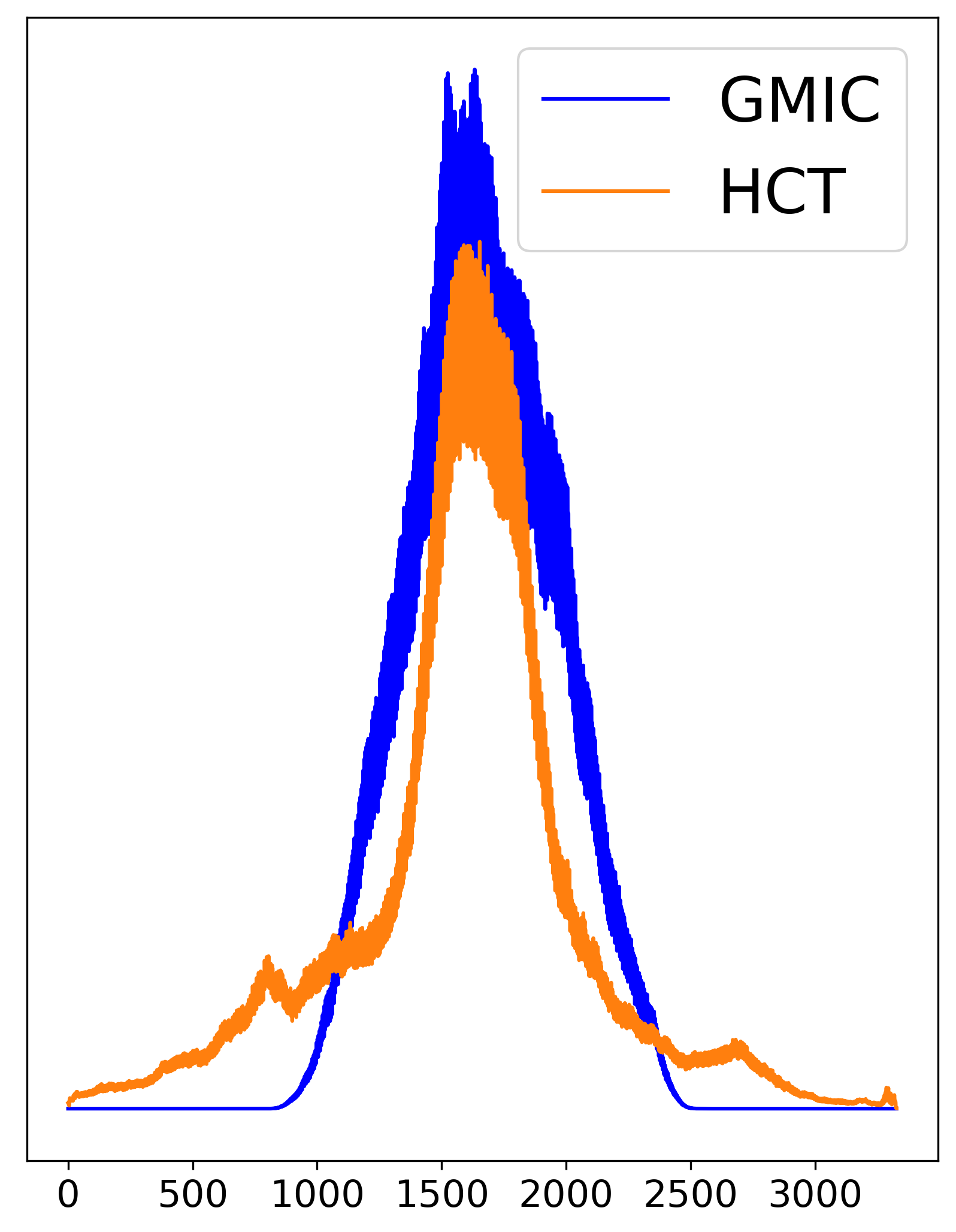} &&
		\includegraphics[width=\wscale\linewidth,height=\hscale\linewidth]{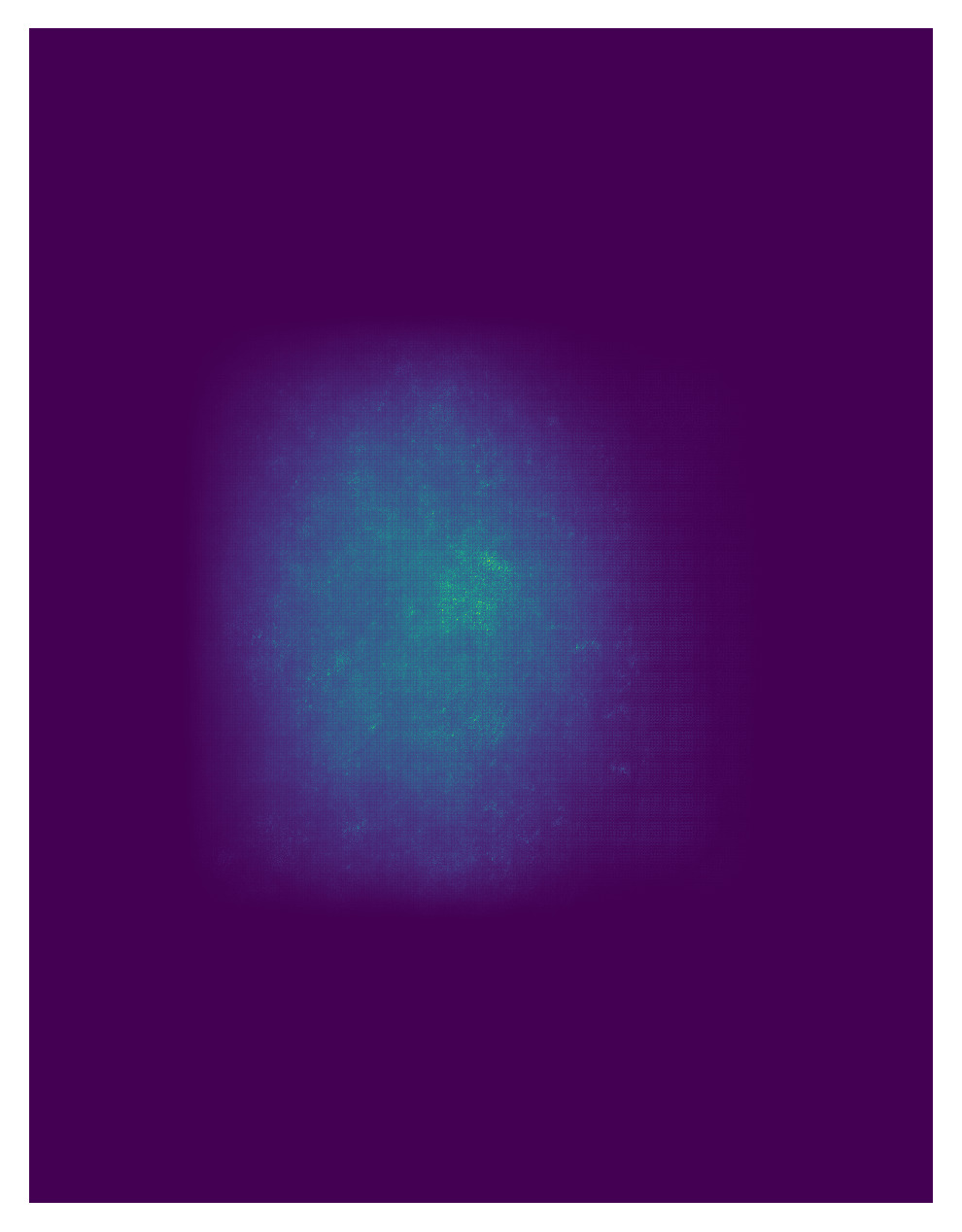} &
		\includegraphics[width=\wscale\linewidth,height=\hscale\linewidth]{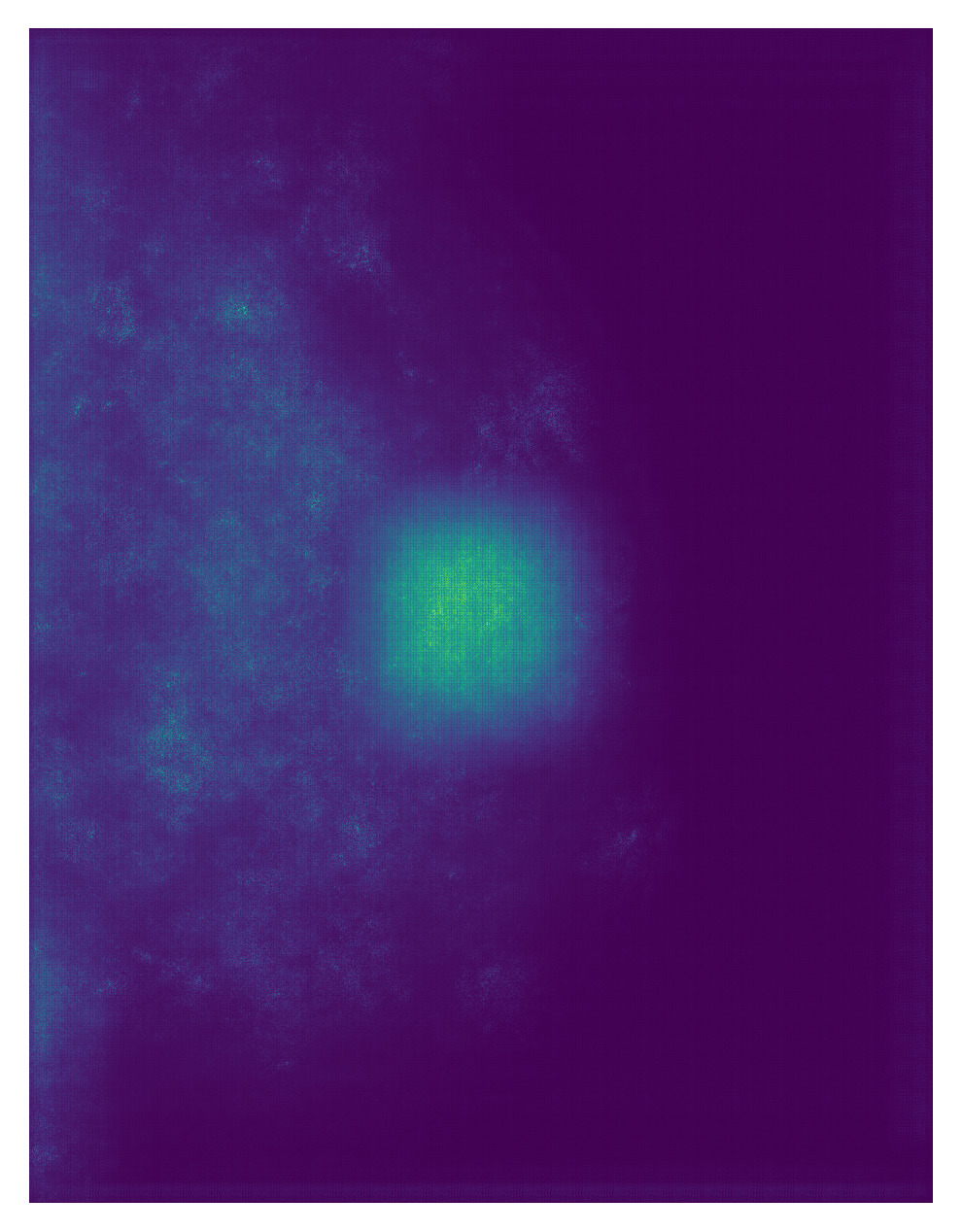} \\
		\bottomrule
	\end{tabular}
\end{table}
\autoref{tab:receptive_field_analysis} compares the rigid ERF of a CNN architecture (\eg GMIC~\cite{shen2019globally}) versus the flexible ERF of HCT.~\autoref{tab:receptive_field_analysis} evaluates the ERF three times (rows): (I) using $N=100$ random breast images (left and right), (II) using $N=100$ random right breast images, and (III) using $N=100$ random left breast images. The GMIC's ERF is rigid and follows a Gaussian distribution. This aligns with Luo~\etal~\cite{luo2016understanding} findings. In contrast, HCT's ERF is dynamic. 

Thanks to the self-attention layers, HCT focuses \textit{dynamically} on regions based on their contents and not their spatial positions. The HCT's ERF is less concentrated on the center but spans both horizontally and vertically according to the image's content. For instance, the second row in~\autoref{tab:receptive_field_analysis} shows how the aggregated ERF of HCT is shifted toward both right (Across Rows) and top (Across Columns) region,~\ie where a right breast is typically located. This emphasizes the value of dynamic attention in transformers.



Transformers are ubiquitous in medical imaging applications. They have been used for classification~\cite{geras2017high,shen2019globally,yang2021momminet,matsoukas2021time,van2021multi}, segmentation~\cite{valanarasu2021medical,gao2021utnet,xie2021cotr,wang2021transbts,zhang2021transfuse,karimi2021convolution}, and image denoising~\cite{zhang2021transct,luthra2021eformer}.  Yet, this recent literature leverages low resolution inputs to avoid the computational cost challenge. Conversely, HCT is designed for high resolution input,~\ie  2-3K pixels per dimension. This is a key difference between our paper and recent literature.  In summary, our key contributions are





\begin{enumerate}[noitemsep]
	\item We propose an efficient high resolution transformer-based model, HCT, for medical images (\autoref{sec:approach}). We demonstrate HCT's superiority using a high resolution mammography dataset (\autoref{sec:experiments}).
	\item We emphasize the importance of dynamic receptive fields for mammography datasets (\autoref{tab:receptive_field_analysis} and \autoref{sec:ablation_study}). This finding is important for medical images where a large portion of an image is irrelevant (\eg an empty region).
\end{enumerate}


\section{HCT: High Resolution Convolutional Transformer}\label{sec:approach}

\begin{figure}[t]
	\centering
	\tiny
	\includegraphics[width=0.9\linewidth]{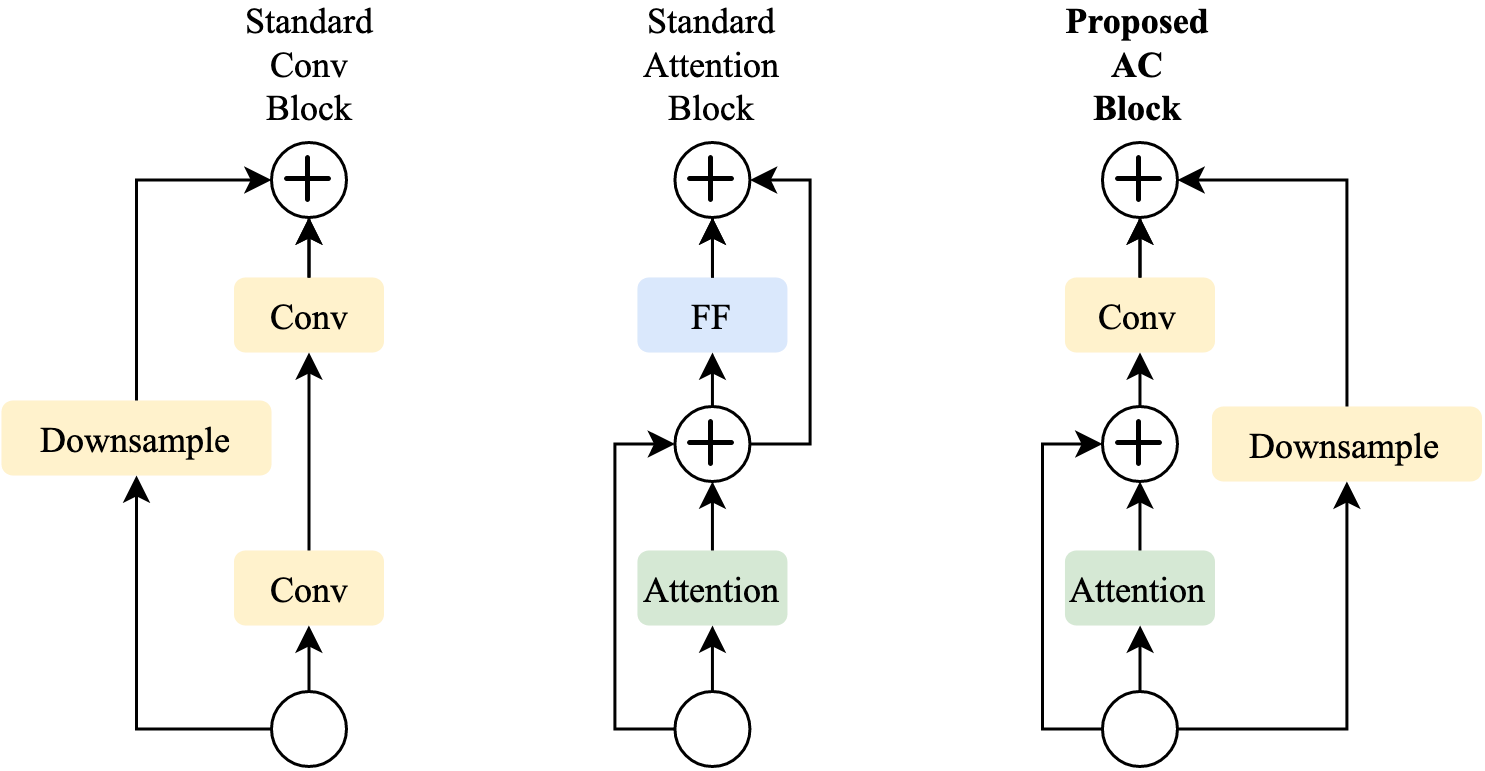}
	\caption{Neural networks' building blocks. (Left) A standard convolutional block for vision models with spatial downsampling capability. (Center) A standard attention block for language models with long range attention capability. (Right) Our \textbf{A}ttention-\textbf{C}onvolutional (AC) block with both spatial downsampling and long range attention capabilities. In the AC block, the conv layer both reduces the spatial resolution  and increases the number of channels. Batchnorm and activation (\eg RELU) layers are omitted for visualization purposes. The AC block's pseudocode is provided in the appendix.}
	\label{fig:ac_block}
\end{figure}

In this section, we propose a network building block that combines both self-attention and convolution operations (\autoref{sec:ac_block}). Then, we leverage a linear self-attention approximation to mitigate the attention layer's quadratic complexity (\autoref{sec:favor}). Finally, we introduce HCT, a convolutional transformer architecture designed for high resolution images (\autoref{sec:hct_arch}).


\subsection{Attention-Convolution (AC) Block}\label{sec:ac_block}
For HCT, we propose an \textbf{A}ttention-\textbf{C}onvolution (AC) building block as shown in~\autoref{fig:ac_block}. The AC block has the following three capabilities: (I) It enables downsampling; this feature is inessential in natural language processing~\cite{bahdanau2014neural,vaswani2017attention} where the number of output and input tokens are equal. Yet, for 2D/3D images, it is important to downsample the spatial resolution as the network's depth increases; (II) The proposed AC block supports different input resolutions. This allows our model to pretrain on small image patches, then finetune on high resolution images. Furthermore, the AC block supports various input resolutions during inference; (III) The AC block consumes and produces spatial feature maps. Thus, it integrates seamlessly into architectures designed for classification~\cite{geras2017high,shen2019globally,shen2020interpretable,matsoukas2021time,yang2021momminet}, detection~\cite{shen2020interpretable,lotter2021robust}, and segmentation~\cite{valanarasu2021medical,gao2021utnet,xie2021cotr,wang2021transbts,zhang2021transfuse,karimi2021convolution}.


To capitalize on these three capabilities, it is vital to reduce the attention layer's computational cost.~\autoref{sec:favor} illustrates how we achieve this. 

\subsection{Efficient Attention}\label{sec:favor}

 \begin{figure}[t]
	\centering
	\scriptsize
	\includegraphics[width=0.8\linewidth]{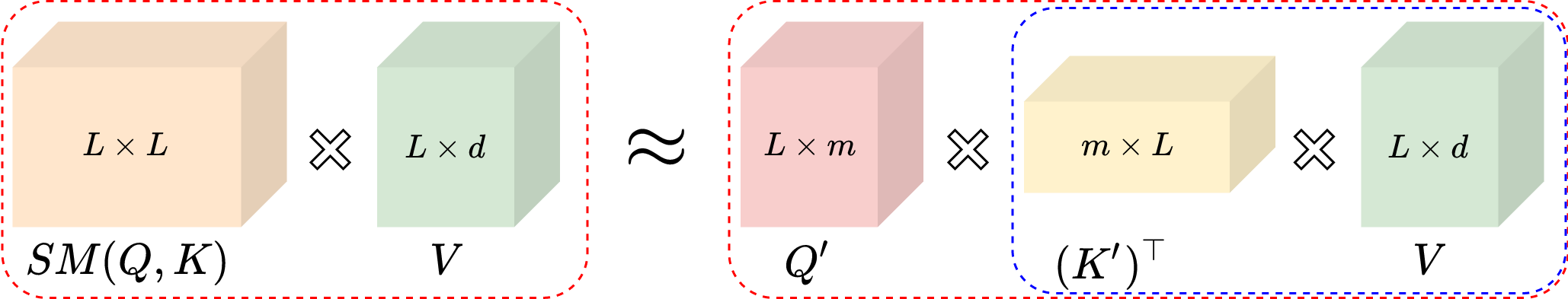}
	\caption{Linear self-attention approximation through kernelization. This reduces the self-attention complexity from quadratic (left) to linear (right). Dashed-blocks indicate the order of computation. SM is the softmax kernel.}
	\label{fig:performer_approx}
\end{figure}

Designing efficient transformers is an active area of research~\cite{tay2020efficient}. To avoid the quadratic complexity of attention layers, we leverage the Fast Attention Via positive Orthogonal Random features approach (FAVOR+), a.k.a. Performers~\cite{choromanski2020rethinking}. In this section, we (1) present an overview of Performers, and (2) compare Performers with other alternatives (\eg Reformer~\cite{kitaev2020reformer}).


Choromanski~\etal~\cite{choromanski2020rethinking} regard the attention mechanism through kernelization,~\ie $\mathbb{K}(x,y)=\phi(x)\phi(y)^\top$, where a kernel $\mathbb{K}$ applied on $x,y \in \mathbb{R}^d$ can be approximated using random feature maps $\phi$ : $\mathbb{R}^d\rightarrow \mathbb{R}_+^m$. To avoid the quadratic complexity ($L\times L$) of $\softmax(Q,K)$, it is rewritten into $\phi(Q)\space\phi(K)^\top = Q'\space(K')^\top$ where $L$ is the number of input tokens, $\softmax$ is the softmax kernel applied on the tokens' queries $Q$ and Keys $K$. Concretely, the softmax kernel is approximated using random feature maps $\phi$ defined as follows
\begin{equation}
	\phi(x) = \frac{h(x)}{\sqrt{m}}\left(f_1(w_1^\top x),...,f_1(w_m^\top x),...,f_l(w_1^\top x), ...,f_l(w_m^\top x)\right),
\end{equation}
where functions $f_1,...,f_l$ : $\mathbb{R}\rightarrow \mathbb{R}$, $h$ : $\mathbb{R}^d\rightarrow \mathbb{R}$, and vectors $w_1,...,w_m \overset{\text{iid}}{\sim}$ $\mathcal{D}$ for some distribution $\mathcal{D}\in \mathcal{P}(\mathbb{R}^d)$.~\autoref{fig:performer_approx} illustrates how Performers change the order of computation to avoid the attention's layer quadratic complexity.

Choromanski~\etal~\cite{choromanski2020rethinking} sampled the vectors $w_1,...,w_m$ from  an isotropic distribution $\mathcal{D}$ and entangled them to be exactly orthogonal. Then, the softmax kernel is approximated using $l=1$, $f_1=\text{exp}$, and $h(x)=\text{exp}(-\frac{||x||^2}{2})$. This formula is denoted as Performer-Softmax. Then, Choromanski~\etal~\cite{choromanski2020rethinking} proposed another stabler random feature maps, Performer-RELU, by setting $l=1$, $f_1=\text{RELU}$, and $h(x)=1$. The simplicity of this kernel approximation is why we choose Performers over other alternatives.


Performers are simpler than other linear approximations. For instance, a Performer leverages standard layers (\eg linear layers). Thus, it is simpler than Reformer~\cite{kitaev2020reformer} which leverages reversible layers~\cite{gomez2017reversible}. A Performer supports global self-attention,~\ie between all input tokens. This reduces the number of hyperparameters compared to block-wise approaches such as shifted window~\cite{liu2021swin}, block-wise~\cite{qiu2019blockwise}, and local-attention~\cite{parmar2018image}. Furthermore, a Performer makes no assumption about the maximum number of tokens while Linformer~\cite{wang2020linformer} does. Accordingly, Performer supports different input resolutions without bells and whistles. Basically, Performers make the least assumptions while being relatively simple. By leveraging Performers in HCT, we promote HCT to other applications beyond our mammography evaluation setup.

 
 \begin{figure}[t]
 	\centering
 	\scriptsize
 	\includegraphics[width=0.7\linewidth]{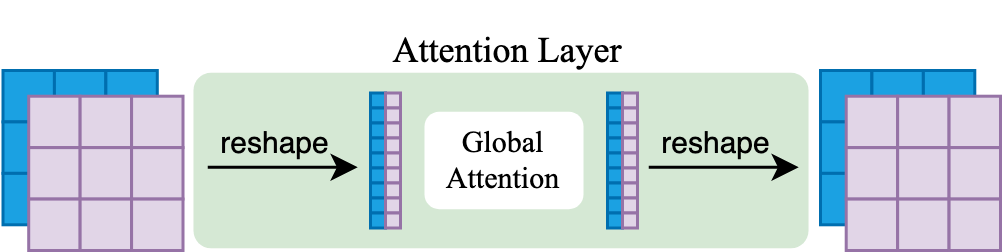}
 	\caption{The proposed self-attention layer flattens the input feature maps before applying global self-attention, then reshapes the result into spatial feature maps. Accordingly, this layer integrates seamlessly in various vision architectures designed for different tasks (\eg classification or detection).}
 	\label{fig:attention_layer}
 \end{figure}

\subsection{The HCT Architecture}\label{sec:hct_arch} 

In this section, we describe the HCT architecture and how the attention layer -- in the AC block -- operates on spatial data,~\eg 2D/3D images. We pick a simple approach that introduces the least hyperparameters. Since self-attention layers are permutation-invariant, we simply flatten the input 3D feature maps from $\mathbb{R}^{W\times H\times C}$ into a sequence of 1D tokens,~\ie $\mathbb{R}^{WH\times C}$ as shown in~\autoref{fig:attention_layer}. This approach supports various input formats (\eg volumetric data) as well as different input resolutions. Furthermore, this approach integrates seamlessly into other vision tasks such as detection and segmentation. Despite its simplicity and merits, the attention layer (\autoref{fig:attention_layer}) brings one last challenge to be tackled.


Attention layers lack inductive bias. Accordingly, transformers are data hungry~\cite{dosovitskiy2020image} while labeled medical images are scarce. Compared to a convolutional layer, the global self-attention layer (\autoref{fig:attention_layer}) is more vulnerable to overfitting and optimization instability. To tackle these challenges, Xiao~\etal~\cite{xiao2021early} have promoted early convolutional layers to increase optimization stability and improve generalization performance. Similar observations have been echoed in recent literature~\cite{cordonnier2019relationship,wu2021cvt,hassani2021escaping,d2021convit}. Accordingly, we integrate our AC block into a variant ResNet-22 architecture. This variant, dubbed GMIC~\cite{shen2019globally}, has been proposed for breast cancer classification. Thus, GMIC serves both as a backbone for architectural changes and as a baseline for quantitative evaluation.

 \begin{figure}[t]
	\centering
	\scriptsize
	\includegraphics[width=0.8\linewidth]{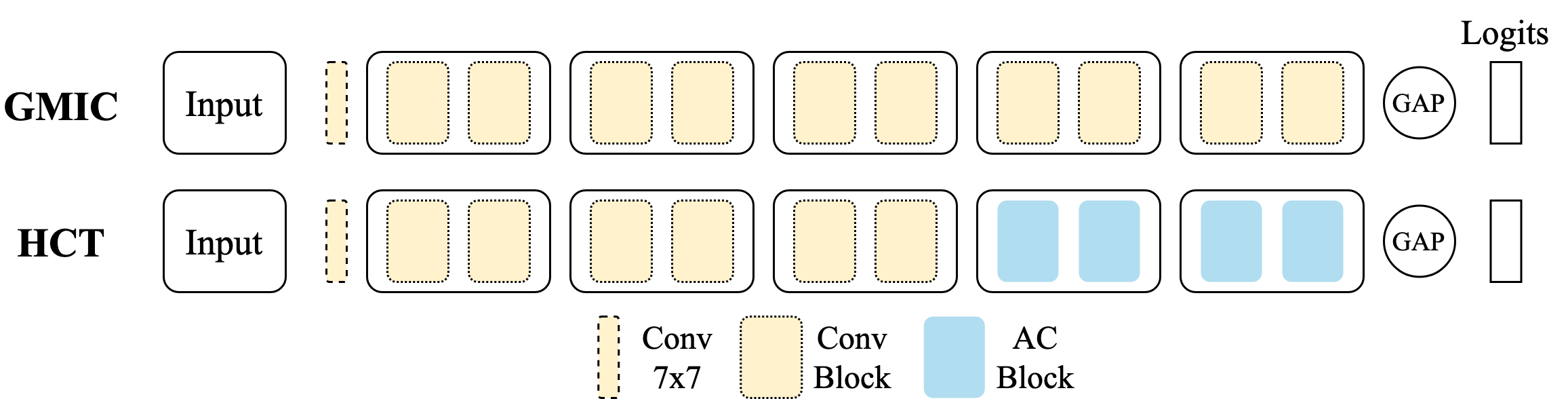}
	\caption{Two mammography classification architectures: GMIC~\cite{shen2019globally} and our HCT. Both architectures are ResNet-22 variants. While GMIC is a pure CNN architecture, HCT is a convolutional transformer-based architecture. GAP denotes global average pooling.}
	\label{fig:archs}
\end{figure}

\autoref{fig:archs} presents both GMIC and HCT architectures, both ResNet-22 variants. Compared to the standard ResNet-22, both architectures employ large strides at early convolutional layers. A large stride (\eg $s=2$) may mitigate but not eliminate the computational cost of high resolution inputs. Both architectures have an initial conv-layer with a $7\times7$ kernel followed by five residual stages. Each stage has two blocks and each block has two layers (\eg conv or attention) as shown in~\autoref{fig:ac_block}. GMIC uses the standard convolutional block~(\autoref{fig:ac_block}, left), while HCT uses the AC block~(\autoref{fig:ac_block}, right). Our experiments~(\autoref{sec:experiments}) explore other architecture variants as baselines for quantitative evaluation.

\section{Experiments}\label{sec:experiments}
\topic{Dataset} 
We evaluate HCT using a high resolution mammography dataset. This dataset comes from the National Health Service OPTIMAM database~\cite{halling2020optimam}. The dataset contains 148,448 full-field digital mammography (FFDM) images from 38,189 exams and 17,322 patients. The dataset is divided randomly into train/val/test splits at the patient level with a ratio of 80:10:10. An image is labeled malignant (positive) if it has a malignant biopsy in the corresponding breast within 12 months of the screening exam. An image is labeled non-malignant (negative) if it has either a benign biopsy within 12 months of the screening exam or 24 months of non-malignant imaging follow-up. Images that violate the above criteria are excluded from the dataset. The validation split is used for hyperparameter tuning.~\autoref{tab:datasets} presents OPTIMAM statistics.


%



\begin{table}[t]
	\centering
	\caption{Statistics of the OPTIMAM mammography dataset.}
	\label{tab:datasets}
	\ra{0.3} 
	\begin{tabular}{@{}lccc@{}}
		\toprule
		~ & Train & Val & Test  \\ \midrule
		Malignant images  (Positive)   & 8,923          & 1,097         & 1,195     \\
		Non-malignant images (Negative) & 49,453          & 6,169         & 6,163     \\
		\midrule
		Total &58,376&7,266& 7,358 \\
		\bottomrule
	\end{tabular}
\end{table}

\begin{table}[t]
	\centering
	\caption{Quantitative evaluation on OPTIMAM. We report both the number of parameters (millions) and the architecture's performance. Performance is reported using AUC and their 95\% confidence intervals (CI) are in square brackets. We evaluate both the small patch and the full image models. The linear-approximation column denotes the linear approximation method employed. E.g., Per-Softmax stands for Performer-Softmax.	The CI is constructed via percentile bootstrapping with 10,000 bootstrap iterations.
	}
	\label{tab:quantitative_evaluation}
	\setlength{\tabcolsep}{3pt}
	\begin{tabular}{@{}lcccc@{}}
		\toprule
		Architecture & Params& Linear Approx. & Patch & Image\\
		\cmidrule{4-5}
		&&& \multicolumn{2}{c}{Adam Optimizer} \\
		\midrule
		GMIC &2.80 &--& 96.13 [95.43, 96.78] & 85.04 [83.74, 86.36] \\
		Def-GMIC & 2.89& -- &96.24 [95.54, 96.88] & 85.17 [83.86, 86.45] \\
		HCT (\textbf{ours}) &1.73& Nyström & \textbf{96.41 [95.76, 97.01]} & 84.83 [83.49, 86.14] \\
		HCT (\textbf{ours}) &1.73& Per-Softmax & 96.35 [95.68, 96.97] & \textbf{86.64 [85.38, 87.86]}\\
		HCT (\textbf{ours}) &1.73 & Per-RELU & 96.34 [95.66, 96.97] & 86.29 [85.02, 87.54] \\
		\midrule
		&&& \multicolumn{2}{c}{Adam + ASAM Optimizer} \\
		\midrule
		GMIC &2.80 &--& 96.29 [95.62, 96.92] & 86.58 [85.34, 87.80]\\
		Def-GMIC & 2.89& -- & 96.71 [96.07, 97.29] & 87.45 [86.24, 88.63] \\
		HCT (\textbf{ours}) &1.73& Nyström & 96.65 [96.02, 97.23] & 86.73 [85.49, 87.95] \\
		HCT (\textbf{ours}) &1.73& Per-Softmax &  96.68 [96.05, 97.26] & 87.39 [86.14, 88.59]\\
		HCT (\textbf{ours}) &1.73 & Per-RELU & \textbf{96.73 [96.09, 97.32]} &  \textbf{88.00 [86.80, 89.18]}\\
		\bottomrule
	\end{tabular}
\end{table}

\medskip
\topic{Baselines} We evaluate HCT using two Performer variants: \textbf{Performer-RELU} and \textbf{Performer-Softmax}. We also evaluate HCT with \textbf{Nyströmformer}~\cite{xiong2021nystr},~\ie another linear attention approximation. The Nyströmformer has a hyperparameter $q$, the number of landmarks. We set $q=\max(W, H)$, where $W$ and $H$ are the width and height of the feature map, respectively. Finally, we evaluate HCT against two CNN architectures: \textbf{GMIC} and \textbf{Def-GMIC}. GMIC~\cite{shen2019globally} is an established benchmark~\cite{frazer2021evaluation} for high resolution mammography. To evaluate all architectures on a single benchmark, GMIC denotes the first module \textit{only} from Shen~\etal~\cite{shen2019globally}. This eliminates any post-processing steps,~\eg extracting ROI proposals. 

We also propose Def-GMIC -- a variant of GMIC -- that uses  deformable~\cite{dai2017deformable} convolutions instead of the attention layer (\autoref{fig:ac_block}, right). Deformable convolutions~\cite{dai2017deformable} introduce a dynamic receptive field and are therefore more flexible compared to a dilated (atrous) convolution~\cite{yu2015multi}. We use these deformable-convolution blocks at the last two layers,~\ie similar to HCT with AC blocks.



%




\medskip
\topic{Technical Details} All networks are initialized randomly, pretrained on small patches ($512\times512$), then finetuned on high resolution images ($3328\times2560$). This is a common practice in mammography literature~\cite{lotter2021robust,frazer2021evaluation}. The patch model is trained using positive and negative patches. Positive patches are centered on bounding box annotations of malignant findings. In contrast, negative patches are randomly selected from within the breast area in non-malignant images.

All models are trained with the cross entropy loss. For the patch model, we use a cosine learning rate decay~\cite{loshchilov2016sgdr} with an initial learning rate $lr=6e^{-4}$, batch size $b=160$, and $e=80$ epochs.  For the image model, we also use the cosine learning rate decay, but with $lr=2e^{-5}$, $b=32$, and $e=80$. All patch models are trained on a single 2080Ti GPU while image models are trained on a single A6000 GPU. 


We evaluate HCT using both Adam~\cite{kingma2014adam} and the adaptive sharpness-aware minimization~\cite{foret2020sharpness,kwon2021asam} (ASAM) optimizers. ASAM simultaneously minimizes the loss value and the loss sharpness. This improves generalization~\cite{foret2020sharpness} especially for transformer-based models~\cite{chen2021vision}. We set the ASAM hyperparameter $\rho=0.05$. We used a public implementation of Performer\footnote{\href{https://github.com/lucidrains/performer-pytorch}{https://github.com/lucidrains/performer-pytorch}}. We use the default Performer hyperparameters~\cite{choromanski2020rethinking},~\ie $n_h=8$ heads, $p=0.1$ dropout rate, $m=\frac{d}{n_h} \log{\frac{d}{n_h}}$, where $d$ is the number of channels. We used standard augmentation techniques (\eg random flipping and Gaussian noise).


\medskip
\topic{Results}~\autoref{tab:quantitative_evaluation} presents a classification evaluation on OPTIMAM. We report both the area under the receiver operating characteristic curve (AUC) and the 95\% confidence interval. With ASAM, Def-GMIC achieves a competitive result, but not the best performance. Furthermore, the Def-GMIC has a Gaussian ERF similar to GMIC; this analysis is reported in the paper appendix. HCT with Performers outperforms GMIC consistently. RELU has been both promoted empirically by Choromanski~\etal~\cite{choromanski2020rethinking} and defended theoretically by Schlag~\etal~\cite{schlag2021linear}. Similarly, we found Performer-RELU superior to Performer-Softmax when the ASAM optimizer was used. With the vanilla Adam optimizer, we attribute Performer-RELU's second best -- and not the best -- performance to stochastic training noise. 



\section{Ablation Study}\label{sec:ablation_study}

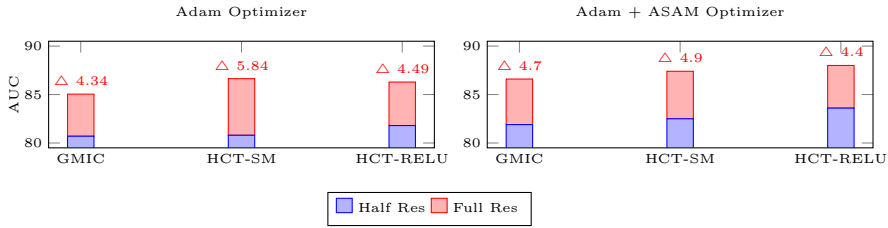
\begin{figure*}[t]
	\centering
	\tiny
	\begin{tikzpicture}
		\begin{groupplot}[group style = {group size = 2 by 1, horizontal sep = 20pt}, 
			ybar stacked,
			height=3.0cm,
			width=0.55\textwidth,
			symbolic x coords={GMIC,HCT-SM,HCT-RELU},
			xtick=data,
			nodes near coords,
			nodes near coords align={vertical},
			x label style={at={(axis description cs:0.5,-0.05)},anchor=north}
			]
			
			\nextgroupplot[title=Adam Optimizer,
			ylabel=AUC,
			ymax=90.5,
			ymin=79.5,
			y label style={at={(axis description cs:0.15,.5)}},
			legend style = { legend columns = -1, legend to name = grouplegend,}
			]	
			
			\addplot+[point meta=explicit symbolic] coordinates {
				(GMIC,80.7) (HCT-SM,80.8) (HCT-RELU,81.8)
			};\addlegendentry{Half Res}
			\addplot+[point meta=explicit symbolic] coordinates {
				(GMIC,\fpeval{85.04-80.7}) [$\triangle$ \fpeval{85.04-80.7}]
				(HCT-SM,\fpeval{86.64-80.8}) [$\triangle$ \fpeval{86.64-80.8}]
				(HCT-RELU,\fpeval{86.29-81.8})  [$\triangle$ \fpeval{86.29-81.8}]
			};\addlegendentry{Full Res}

			\nextgroupplot[title=Adam + ASAM Optimizer,
			ymax=90.5,
			ymin=79.5,
			y label style={at={(axis description cs:0.15,.5)}},
			legend style = { legend columns = -1, legend to name = grouplegend,}
			]	
			\addplot+[point meta=explicit symbolic] coordinates {
				(GMIC,81.9) (HCT-SM,82.5) (HCT-RELU,83.6)
			};\addlegendentry{Half Res}
			\addplot+[point meta=explicit symbolic] coordinates {
				(GMIC,\fpeval{86.6-81.9}) [$\triangle$ \fpeval{86.6-81.9}]
				(HCT-SM,\fpeval{87.4-82.5}) [$\triangle$ \fpeval{87.4-82.5}]
				(HCT-RELU,\fpeval{88.0-83.6})  [$\triangle$ \fpeval{88.0-83.6}]
			};\addlegendentry{Full Res}
			
		\end{groupplot}
		\node[below] at ($(group c1r1.south) +(2.5,-0.5)$) {\pgfplotslegendfromname{grouplegend}}; 
	\end{tikzpicture}
	
	\caption{Quantitative evaluation using half ($1664\times1280$) and full ($3328\times2560$) resolution inputs. $\triangle$ denotes the \textit{absolute} improvement margin achieved on the full resolution by the same architecture. This highlights the advantage of high resolution inputs when looking for malignant tissues at an early stage.}
	
	\label{fig:half_res_evaluation}
\end{figure*}

In this section, we compute the effective receptive field (ERF) of both GMIC and HCT. Through this qualitative evaluation, we demonstrate HCT's fitness for medical images. Finally, we evaluate the performance of HCT using half and full resolution inputs.

\medskip
\topic{Effective Receptive Field Study} We follow Luo~\etal~\cite{luo2016understanding} procedure to compute the ERF. Specifically, we feed an input image to our model. Then, we set the gradient signal to one at the center pixel of the output feature map, and to zero everywhere else. Then, we back-propagate this gradient through the network to get the input's gradients. We repeat this process using $N=100$ randomly sampled breast images. Finally, we set the ERF to the mean of these $N$ inputs' gradients.~\autoref{tab:receptive_field_analysis} presents the ERF of both GMIC and HCT with Performer-RELU. HCT is superior in terms of modeling long range dependencies.

\medskip
\topic{Resolution Study} To emphasize the importance of high resolution inputs for medical images, we train GMIC and HCT using both half and full resolutions inputs.~\autoref{fig:half_res_evaluation} delivers an evaluation of all models: GMIC, HCT with Performance-RELU (HCT-RELU), and HCT with Performer-Softmax (HCT-SM). The full resolution models beat the half resolution models with a significant margin (+4\% AUC).

\medskip
\topic{Discussion} Compared to a low resolution input, a high resolution input will always pose a computational challenge. As technology develops, we will have better resources (\eg GPUs), but image-acquisition quality will improve as well. HCT is a ResNet-22 variant, a shallow architecture compared to ResNet-101/152 and recent architectural development~\cite{coates2013deep,brown2020language,yuan2021florence}. For high resolution inputs, a deep network is a luxury.


\section{Conclusion}\label{sec:conclusion}
We have proposed HCT, a  convolutional transformer for high resolution inputs. HCT leverages linear attention approximation to avoid the quadratic complexity of vanilla self-attention layers. Through this approximation, HCT boosts performance, models long range dependencies, and maintains a superior effective receptive field.

%
%
%
\bibliographystyle{splncs04}
\bibliography{hct}

\clearpage
\section{Appendix}

\newcommand{\beginsupplement}{%
	\setcounter{table}{0}
	\renewcommand{\thetable}{A\arabic{table}}%
	\setcounter{figure}{0}
	\renewcommand{\thefigure}{A\arabic{figure}}%
	\setcounter{section}{0}
	\renewcommand{\thesection}{A\arabic{section}}%
	\setcounter{equation}{0}
	\renewcommand{\theequation}{A\arabic{equation}}%
}

\beginsupplement

In this section, we present further quantitative and qualitative evaluations for HCT. Algorithm~\autoref{alg:ac_block} presents a pseudo code for the proposed AC block. A complete implementation is available on Github\footnote{\href{https://bit.ly/3ykBhhf}{https://bit.ly/3ykBhhf}}.



\begin{algorithm}[t]
	\tiny
	\caption{The proposed AC block's forward pass. This implementation follows GMIC's implementation \href{https://github.com/nyukat/GMIC/blob/e8f539f2515f648cdd3a7c624363abdd3f4f973a/src/modeling/modules.py\#L51}{src/modeling/modules.py\#L51}.}\label{alg:ac_block}
	\tiny
	\begin{tabular}{@{}cc@{}}
		\begin{lstlisting}[language=Python]
def forward(self, x):
	residual = x
	
	out = self.bn1(x)
	out = self.relu(out)
	if self.downsample is not None:
		residual = self.downsample(out)		
	out = self.linear_attention(out)
	
	out += x
	
	out = self.bn2(out)
	out = self.relu(out)
	# This conv controls both
	# resolution and # of channels
	out = self.conv2(out)
	out += residual
	return out\end{lstlisting}
		&
		\begin{lstlisting}[language=Python]
def linear_attention(self, x):
	bz, num_chns, rows, cols = x.shape
	
	## Reshape for Transformer
	x = x.view(bz, num_chns, -1) 
	x = x.permute(0, 2, 1)
	
	# We recommend Performer-RELU. Yet,
	# other linear approximations work
	q = self.attention(x)
	
	## Reshape for CNNs
	q = q.permute(0, 2, 1)
	q = q.view(bz, -1, rows, cols)  
	return q
			
			
		\end{lstlisting}
	\end{tabular}
\end{algorithm}

\begin{figure}[t]
	\centering
	\tiny
	\begin{subfigure}[t]{0.48\textwidth}
		\centering
		\includegraphics[width=0.65\textwidth,height=0.65\textwidth]{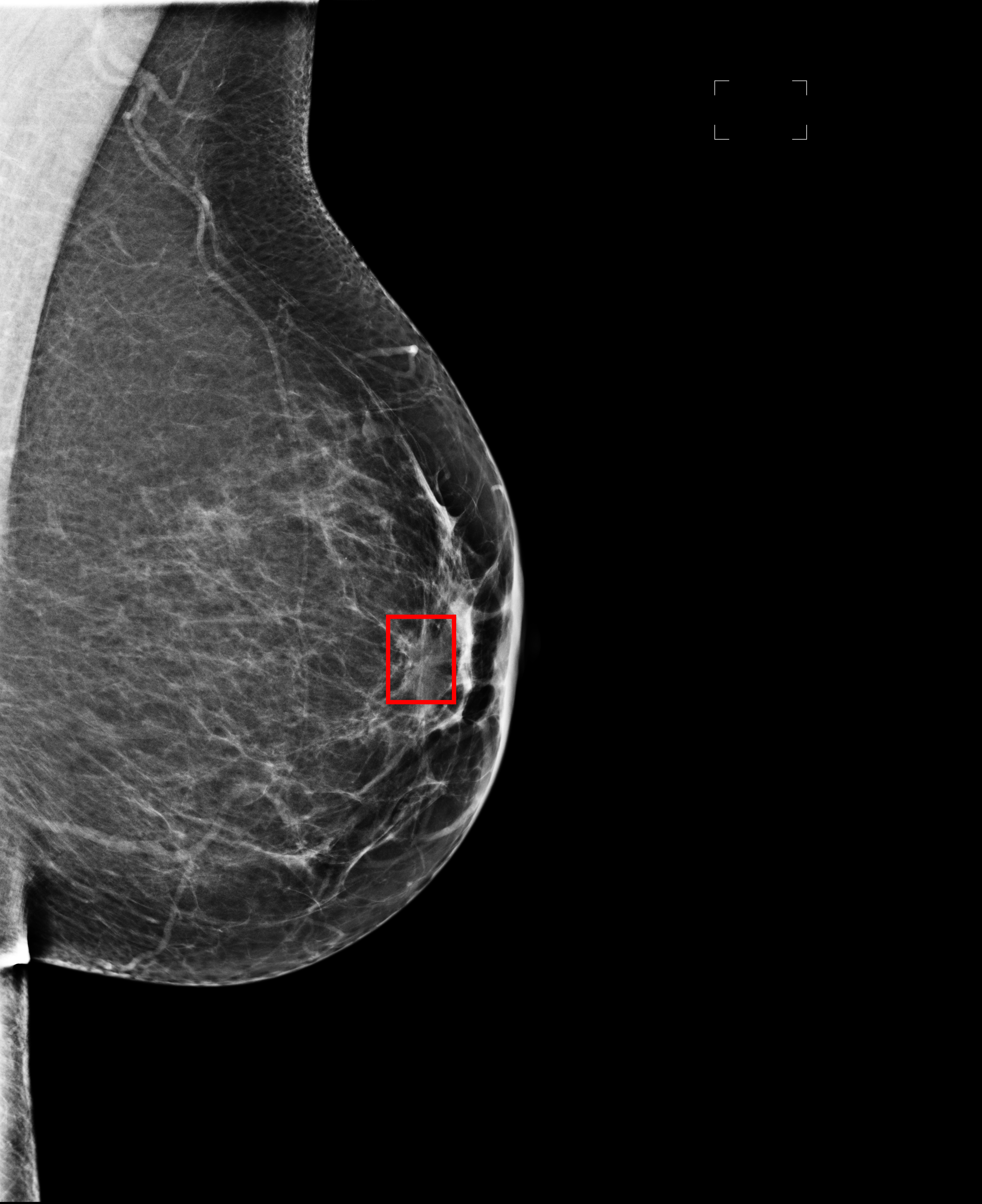}
		\label{fig:finding_mammograms_a}
	\end{subfigure}\hfill\begin{subfigure}[t]{0.48\textwidth}
		\centering
		\includegraphics[width=0.65\textwidth,height=0.65\textwidth]{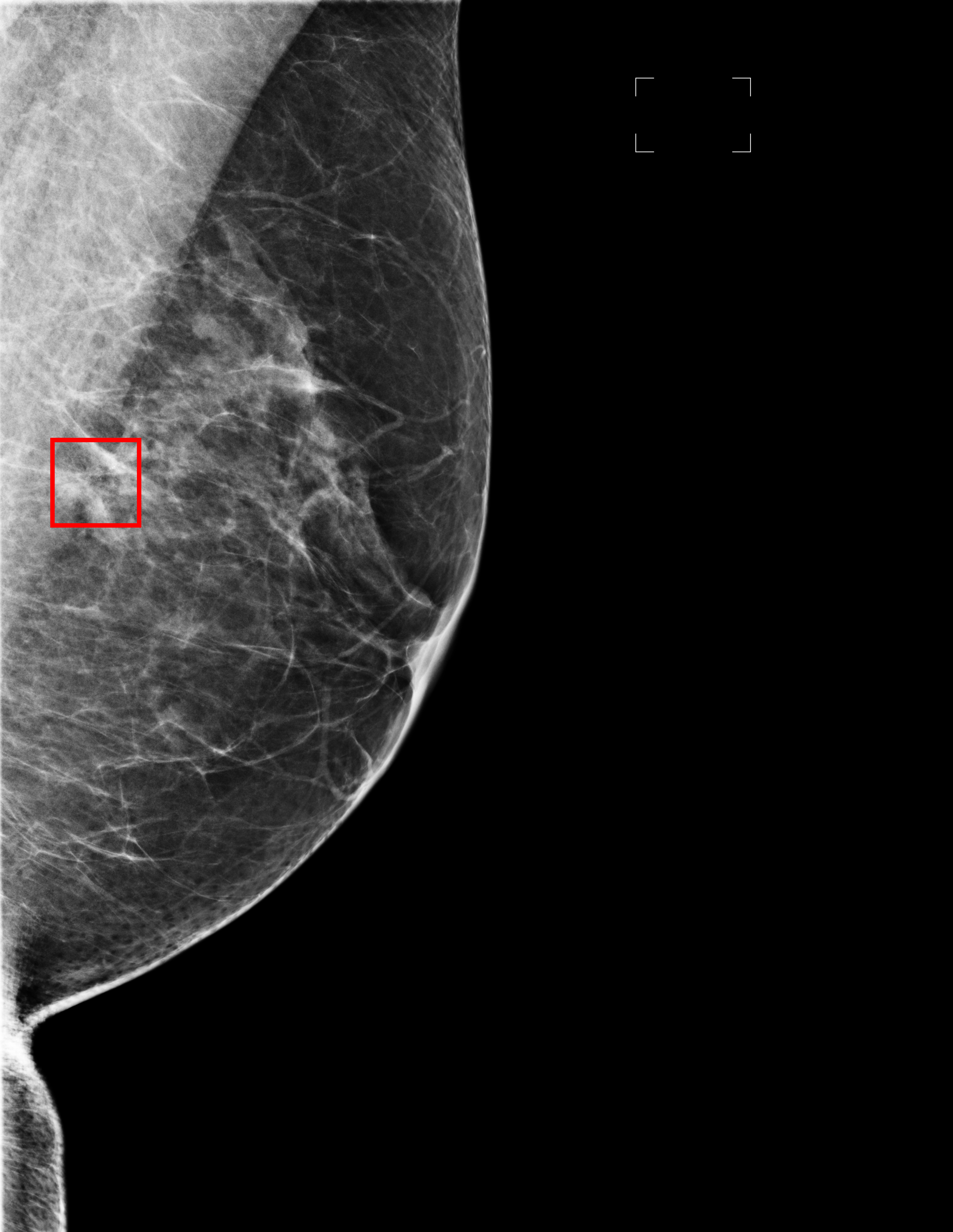}
		\label{fig:finding_mammograms_b}
	\end{subfigure}
	\caption{Different malignant findings. (Left) Architectural Distortion: Having a greater receptive field aids with this finding because the tissue lines follow a different pattern than the pattern of the tissues in its greater surrounding area. (Right) Focal Asymmetry: These are density findings that do not fit the criteria of a mass; it lacks a definition like a convex outer border. So, a greater global vision is beneficial for classifying these cases.}
	\label{fig:finding_mammograms}
\end{figure}

\autoref{tab:finding_types} breaks down the models' performance on four finding types: (a) Mass --- an area of breast tissue defined with abnormal shape and edges, 
(b) Calcifications --- tiny calcium deposits within the breast tissue, (c) Focal Asymmetry --- density findings seen on two views but lack definition, and (d) Architectural Distortion (arch. distortion) --- an area of distorted breast tissues. For each finding type, the positive class includes all malignant images with this finding type, while the negative class includes all negative images,~\ie no findings at all. For each finding type, HCT-RELU performs the best. The largest performance gains are achieved on the architectural distortion (+5.0\%) and the focal asymmetry (+4.5\%) finding types. \autoref{fig:finding_mammograms} depicts examples of these finding types and explains how a larger effective receptive field is helpful given the findings' characteristics.


\begin{table}[t]
	\centering
	\caption{Quantitative classification evaluation using different finding types. All models were trained with ASAM optimizer. Performance is reported using the AUC on image level.}
	\label{tab:finding_types}
	\setlength{\tabcolsep}{3pt}
	\begin{tabular}{@{}lccccc@{}}
		\toprule
		Architecture & Linear Approx. & Mass & Calcifications & Arch. Distortion & Focal Asymmetry \\
		\midrule
		GMIC  &--& 90.1 & 91.5 & 78.1 & 80.5 \\
		Def-GMIC & -- & 91.2 & 91.6 & 82.5 & 83.7 \\
		HCT (\textbf{ours}) & Nyström & 90.2 & 90.9 & 81.0 & 83.1 \\
		HCT (\textbf{ours}) & Per-Softmax & 90.9 & 92.2 & 81.2 & 85.3 \\
		HCT (\textbf{ours})  & Per-RELU & \textbf{91.3} & \textbf{92.4} & \textbf{82.6} & \textbf{85.5} \\
		\bottomrule
	\end{tabular}
\end{table}

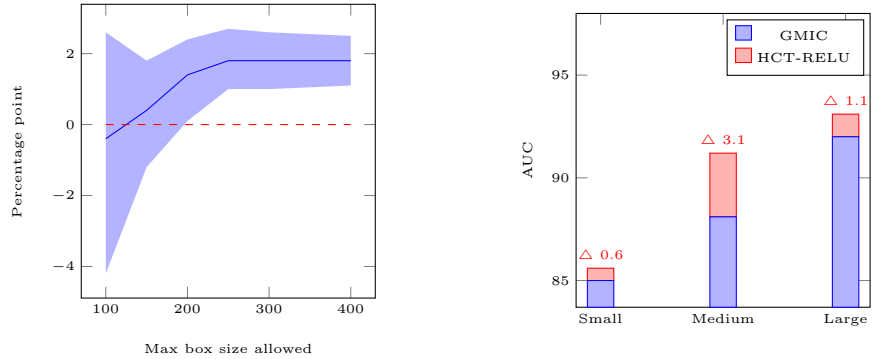
\begin{figure*}[t]
	\begin{subfigure}[t]{0.45\textwidth}
		\centering
		\tiny
		\begin{tikzpicture}
			\begin{axis}[
				height=\textwidth,
				width=\textwidth,
				ylabel={Percentage point},
				xlabel={Max box size allowed},
				y label style={at={(axis description cs:0.1,.5)}},
				]
				\addplot[color=blue,solid] coordinates {
					(100, -0.4)(150,0.4)(200, 1.4)(250, 1.8)(300, 1.8)(400, 1.8)
				};
				\addplot [name path=bluelower,draw=none]coordinates {
					(100, -4.2)(150, -1.2)(200,0.1)(250, 1)(300, 1)(400, 1.1)
				};
				\addplot [name path=blueupper,draw=none]coordinates {
					(100,2.6)(150,1.8)(200,2.4)(250, 2.7)(300,2.6)(400, 2.5)
				};
				\addplot [fill=blue!30] fill between[of=blueupper and bluelower];
				\addplot[red,dashed,update limits=false] 
				coordinates {(100,0) (400,0)} 
				node[above] at (axis cs:300,0) {};
			\end{axis}
		\end{tikzpicture}
		\captionsetup{font=footnotesize}
		\captionof{figure}{AUC difference between HCT-RELU and GMIC. The x-axis denotes a threshold for filtering malignant images. Images with malignant findings smaller than a threshold are considered positive during AUC evaluation. The shaded area represents the 95\% confidence interval of this difference.
			 }
		\label{fig:finding_size_upto}
	\end{subfigure}\hfill\begin{subfigure}[t]{0.45\textwidth}
		\centering
		\tiny
		\begin{tikzpicture}
			\begin{axis}[ 
				ybar stacked,
				height=\textwidth,
				width=\textwidth,
				xtick=data,
				xlabel={\space},
				nodes near coords,
				nodes near coords align={vertical},
				symbolic x coords={Small, Medium, Large},
				ylabel=AUC,
				ymax=98,
				y label style={at={(axis description cs:0.15,.5)}},
				]
				\addplot+[point meta=explicit symbolic] coordinates {
					(Small,85.0) (Medium,88.1) (Large,92.0)
				};\addlegendentry{GMIC}
				\addplot+[point meta=explicit symbolic] coordinates {
					(Small,\fpeval{85.6-85.0}) [$\triangle$ \fpeval{85.6-85.0}]
					(Medium,\fpeval{91.2-88.1}) [$\triangle$ \fpeval{91.2-88.1}]
					(Large,\fpeval{93.1-92.0})  [$\triangle$ \fpeval{93.1-92.0}]
				};\addlegendentry{HCT-RELU}
			\end{axis}
		\end{tikzpicture}
		\captionsetup{font=footnotesize}
		\captionof{figure}{Model performance for different malignant finding sizes. $\triangle$ denotes the \textit{absolute} improvement margin of HCT-RELU over GMIC.}
		\label{fig:finding_size_category}
	\end{subfigure}
	\caption{Quantitative evaluation of HCT-RELU and GMIC using malignant-findings' sizes.}
	\label{fig:finding_size}
\end{figure*}


 In our test split, we have 1195 malignant images. Out of these images, 1004 images have bounding box annotations that outline the malignant tissue. The square root of the bounding box's area  -- width times height -- denotes the finding's size. \autoref{fig:finding_size} ablates the models' performance through these findings' sizes.


~\autoref{fig:finding_size_upto} evaluates both GMIC (dashed red) and HCT-RELU (solid blue) using findings' sizes. The x-axis denotes seven finding-size thresholds \{100, 150, 200, 250, 300, 350, 400\} used to filter malignant images. For each threshold, images with findings smaller than the threshold are considered positive. Thus, these thresholds define positive images for AUC evaluation against all negative images. The y-axis denotes the AUC difference between HCT-RELU and GMIC. For small findings ($< 100$), the AUC difference is marginal. As the finding-size threshold increases (\eg $> 200$), more positive images are included and HCT-RELU outperforms GMIC significantly. HCT-RELU's 95\% confidence interval lies above the GMIC baseline.

~\autoref{fig:finding_size_category} evaluates both GMIC and HCT-RELU on three subsets. The small, medium and large subsets contain positive (malignant) images with the bottom, middle and top $\frac{1}{3}$ finding sizes, respectively. Every positive subset is evaluated against all negative images. HCT-RELU outperforms GMIC on all subsets. Both HCT-RELU and GMIC suffer on small findings but excel on large findings. Compared to GMIC, HCT-RELU achieves a marginal absolute improvement of 0.6\% on the small subset but a significant 3.1\% on the medium subset. Even on the large subset where GMIC achieves its best performance, HCT-RELU still boosts performance by an absolute 1.1\%.

\autoref{tab:receptive_field_analysis_def_gmic} compares the ERFs of CNN architectures (GMIC and Def-GMIC) versus the ERF of HCT. This table shows the aggregated ERF only-- across columns ($1^{\text{st}}$ row) and rows ($2^{\text{nd}}$ row) -- for brevity. Both GMIC and Def-GMIC have ERFs that follow a Gaussian distribution independent of the input signal. Conversely, our HCT's ERF is dynamic and depends on the input signal.

\newcommand{\ahscale}{0.24} 
\newcommand{\awscale}{0.20} 
\begin{table}[t]
	\scriptsize
	\centering
	\caption{Effective receptive field (ERF) analysis for GMIC (Blue), Def-GMIC (Green), and HCT (Orange). These ERFs are generated using $N=100$ randomly sampled left and right breasts ($1^{\text{st}}$ column), right breasts ($2^{\text{nd}}$ column), and left breasts ($3^{\text{rd}}$ column). To highlight the ERF difference, we aggregate the ERF across columns ($1^{\text{st}}$ row) and rows ($2^{\text{nd}}$ row). These high resolution images are best viewed on a screen.}
	\label{tab:receptive_field_analysis_def_gmic}
	\setlength{\tabcolsep}{2pt}
	\begin{tabular}{@{}ccc@{}}
		\toprule
		Left and Right & Right& Left\\
		\midrule
		\includegraphics[width=\awscale\linewidth,height=\ahscale\linewidth]{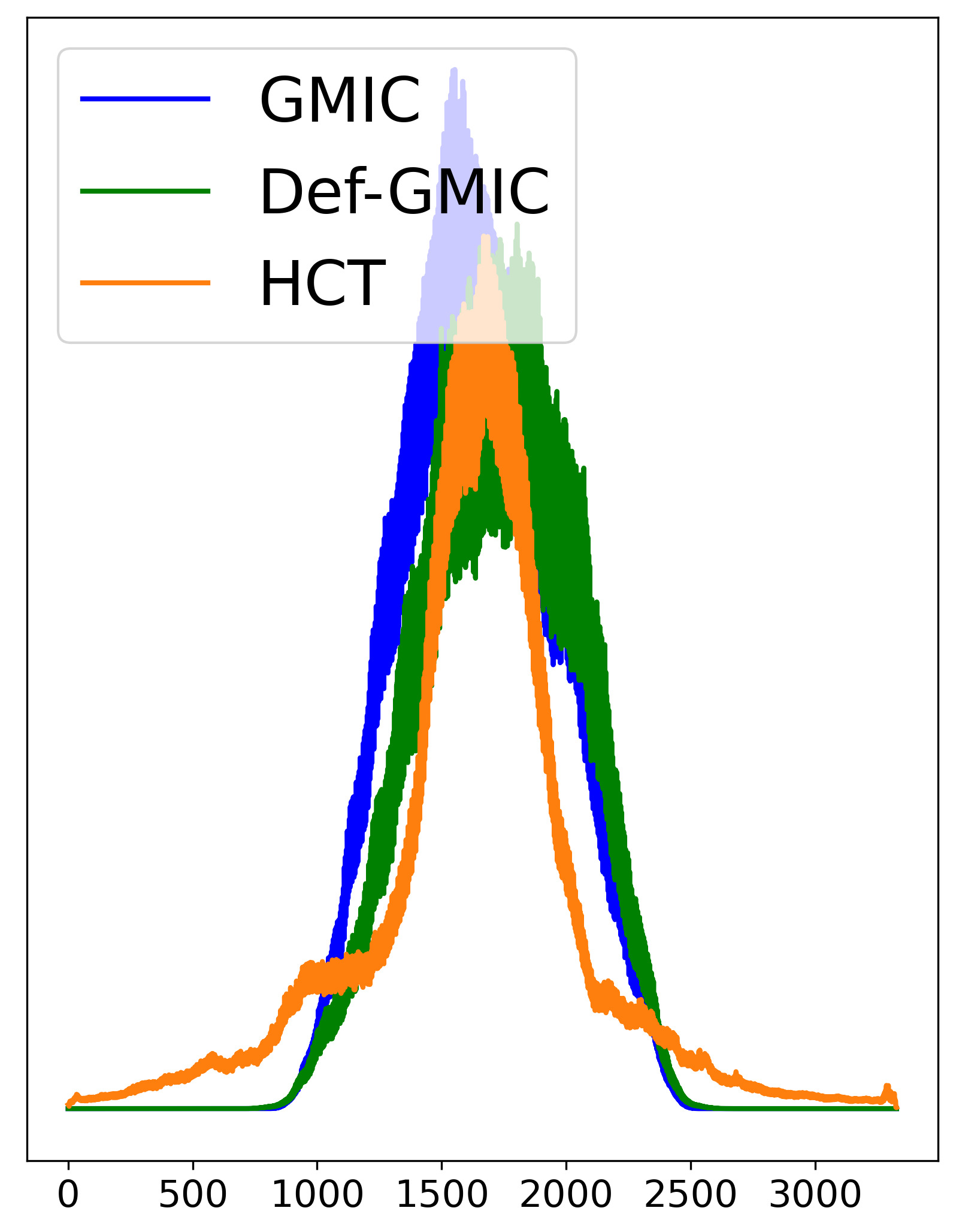} &
		\includegraphics[width=\awscale\linewidth,height=\ahscale\linewidth]{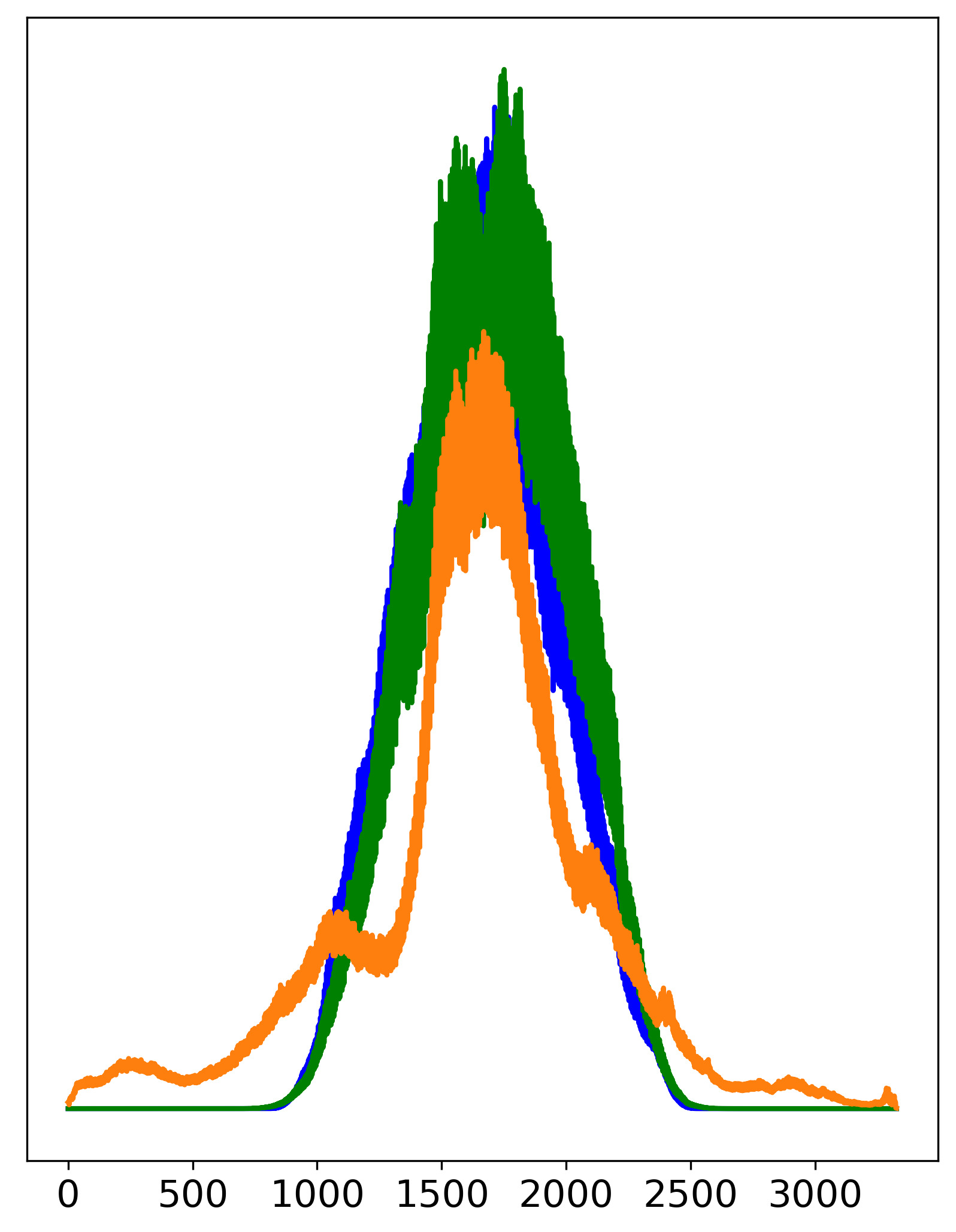} &
		\includegraphics[width=\awscale\linewidth,height=\ahscale\linewidth]{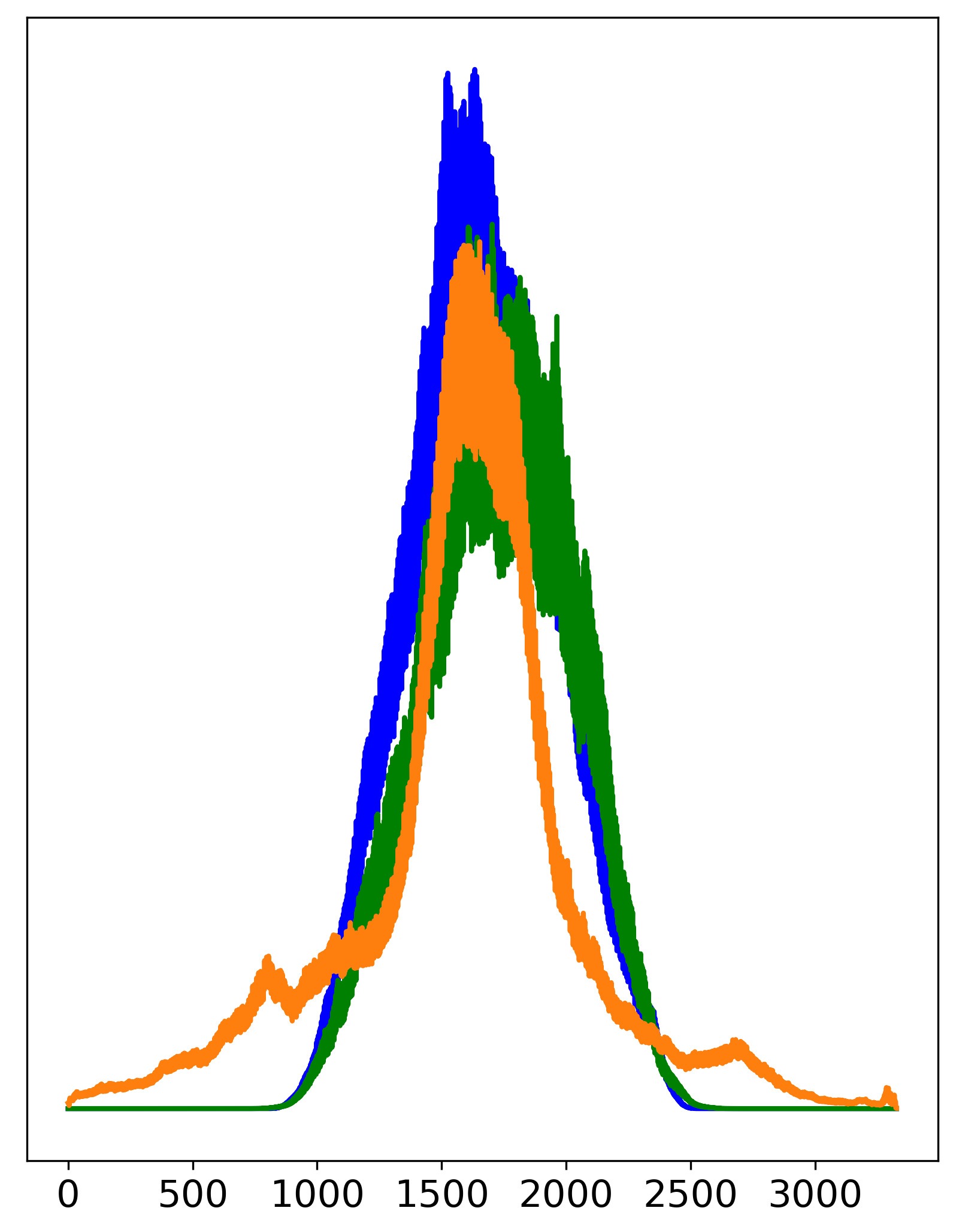}\\
		
		\midrule
		\includegraphics[width=\awscale\linewidth,height=\ahscale\linewidth]{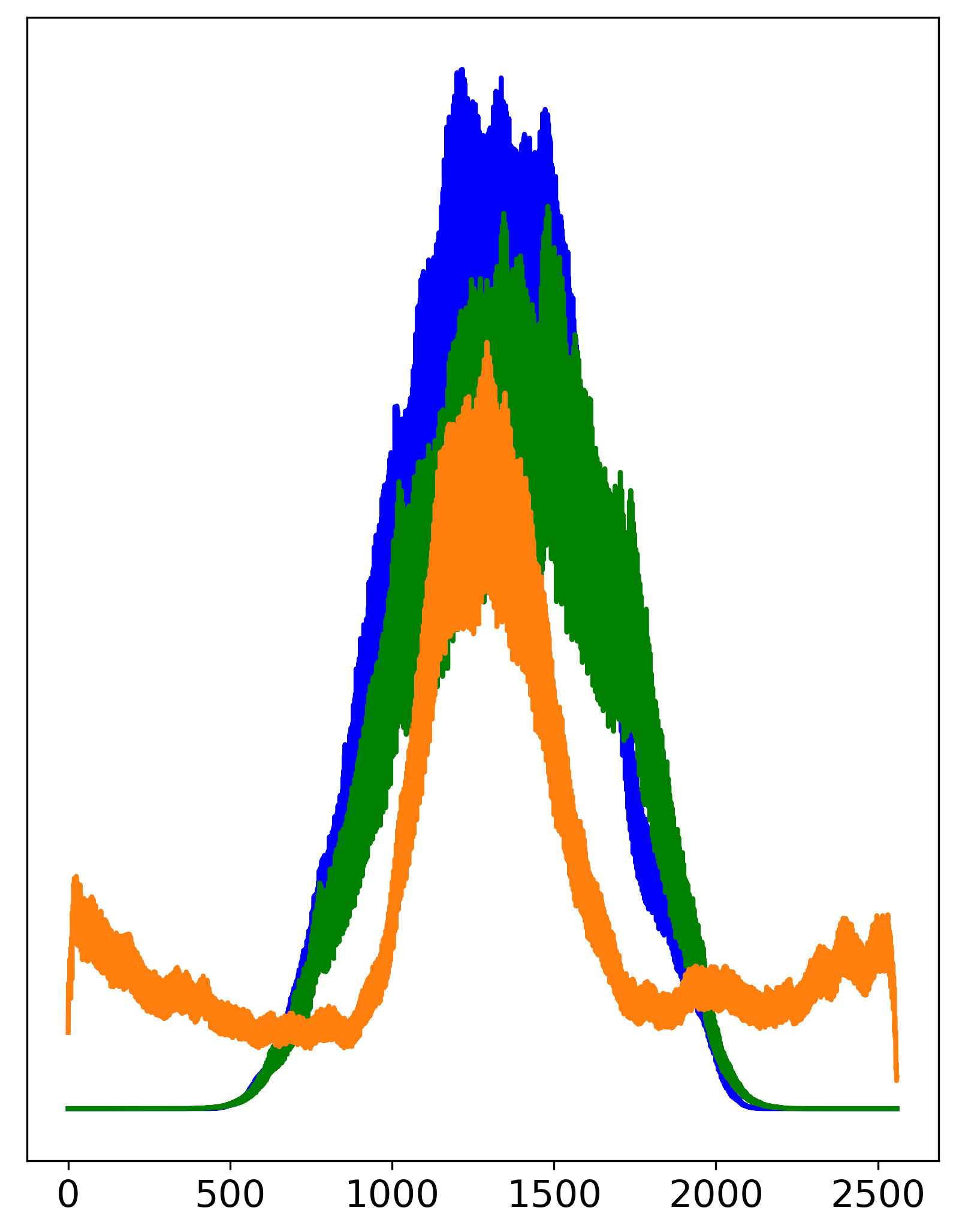} &
		\includegraphics[width=\awscale\linewidth,height=\ahscale\linewidth]{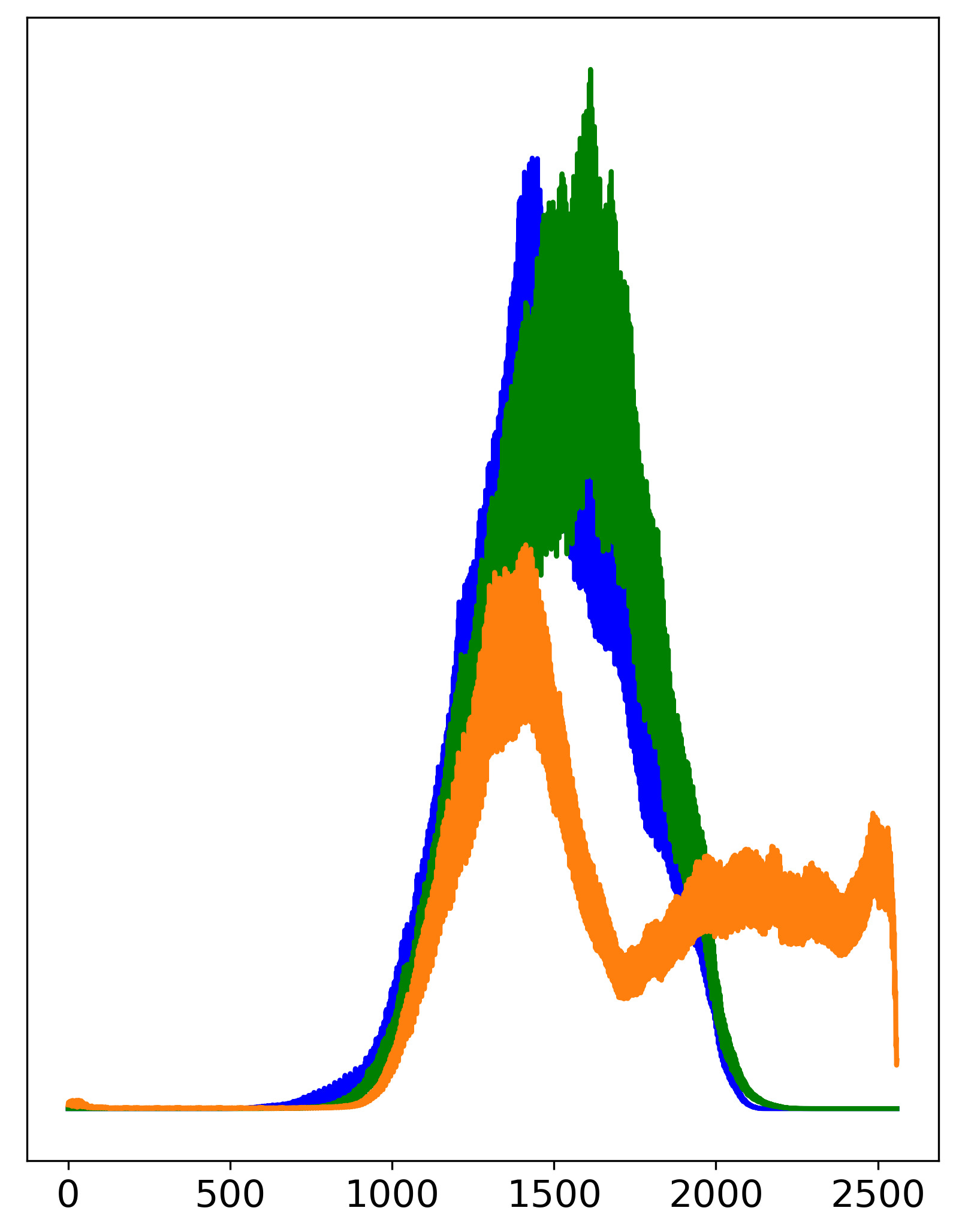} &
		\includegraphics[width=\awscale\linewidth,height=\ahscale\linewidth]{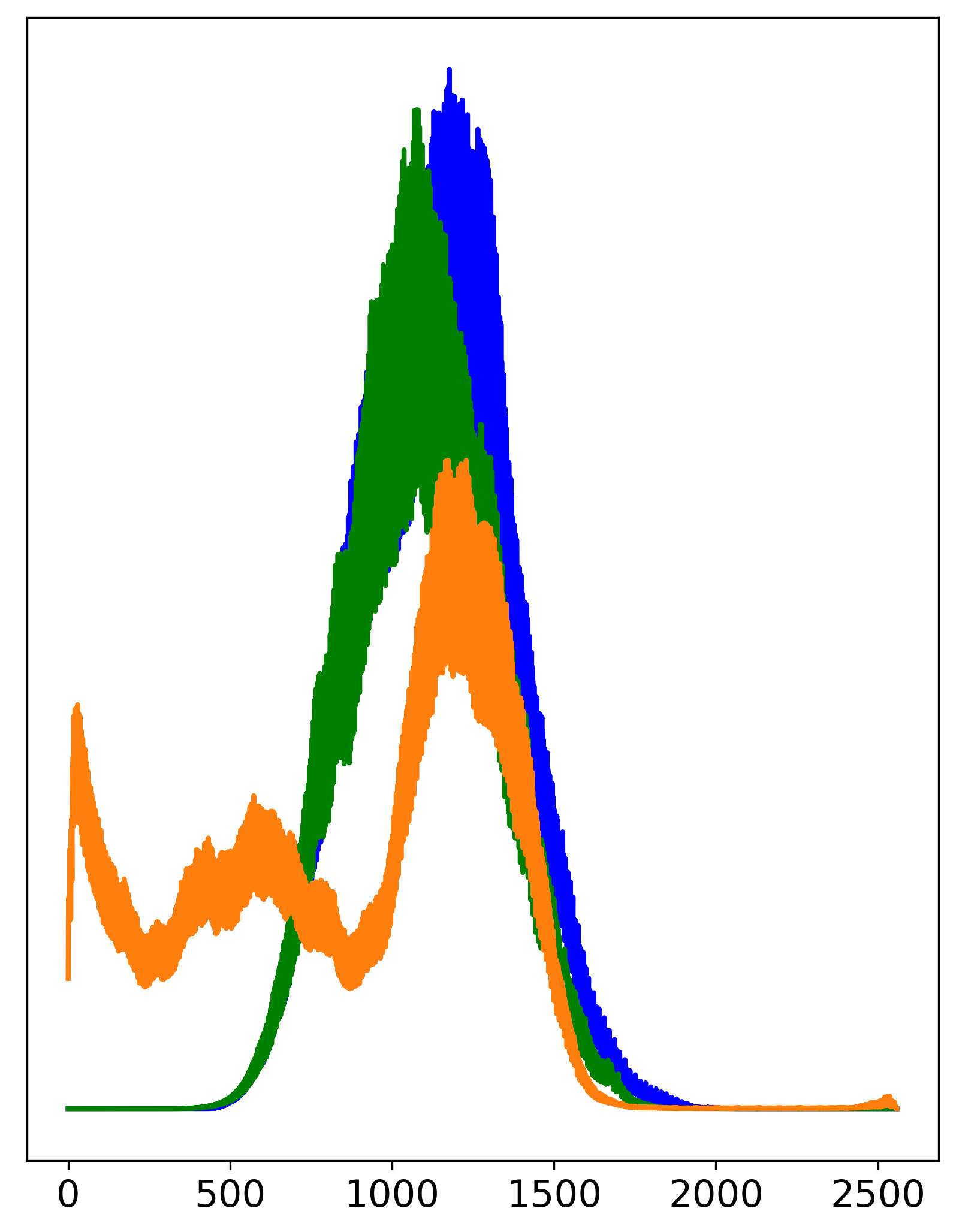}\\
		
		\bottomrule
	\end{tabular}
\end{table}

\end{document}